%% file: main.tex
\documentclass[10pt,twocolumn,letterpaper]{article}

\usepackage[pagenumbers]{cvpr} 

\usepackage[acronym]{glossaries}
\usepackage[linesnumbered,ruled,vlined]{algorithm2e}
\usepackage{tikz}
\usepackage{array}
\usepackage{booktabs}
\usepackage{multirow}
\usepackage{colortbl} 
\usepackage{diagbox} 
\definecolor{lightgray}{gray}{0.9} 
\usepackage{amsthm}

\usepackage{amssymb} 
\usepackage{pifont}  
\usepackage{graphicx} 

\newcommand{\cmark}{\checkmark}

\newacronym{nas}{NAS}{Neural Architecture Search}
\newacronym{llm}{LLM}{Large Language Models}
\newtheorem{theorem}{Theorem}
\newtheorem{definition}{Definition}

\usepackage[accsupp]{axessibility}
\usepackage{fancyhdr}
\usepackage[dvipsnames]{xcolor}

\input{preamble}

\definecolor{cvprblue}{rgb}{0.21,0.49,0.74}
\usepackage[pagebackref,breaklinks,colorlinks,allcolors=cvprblue]{hyperref}

\def\paperID{11379} 
\def\confName{CVPR}
\def\confYear{2025}

\title{L-SWAG: Layer-Sample Wise Activation with Gradients information for Zero-Shot NAS on Vision Transformers\thanks{Published as a conference paper at CVPR 2025}\vspace{-0.05in}}

\author{Sofia Casarin$^1$,
Sergio Escalera$^{2,3}$,
Oswald Lanz$^1$ \\
$^1$Free University of Bozen-Bolzano, Bolzano, Italy \\
$^2$Computer Vision Center, Barcelona, Spain \\
$^3$Universitat de Barcelona, Barcelona, Spain \\
{\tt\small scasarin@unibz.it, sergio@maia.ub.es, lanz@inf.unibz.it}
\and
}

\begin{document}

\maketitle
\input{sec/0_abstract}    
\input{sec/1_intro}

\input{sec/2_related_works}
\input{sec/3_methods}

\input{sec/4_experiments}
\input{sec/5_conclusions}

\input{sec/X_suppl}

{
    \small
    \bibliographystyle{ieeenat_fullname}
    \bibliography{main}
}



\end{document}

%% file: preamble.tex
\input{math_commands}

%% file: math_commands.tex
\usepackage{amsmath,amsfonts,bm}
\usepackage{dsfont}

\def\eqref#1{equation~\ref{#1}}

\def\1{\bm{1}}

\def\va{{\bm{a}}}

\def\vp{{\bm{p}}}

\def\vx{{\bm{x}}}
\def\vy{{\bm{y}}}

\def\mX{{\bm{X}}}

\DeclareMathAlphabet{\mathsfit}{\encodingdefault}{\sfdefault}{m}{sl}
\SetMathAlphabet{\mathsfit}{bold}{\encodingdefault}{\sfdefault}{bx}{n}



%% file: sec/0_abstract.tex
\begin{abstract}
Training-free Neural Architecture Search (NAS) efficiently identifies high-performing neural networks using zero-cost (ZC) proxies.~Unlike multi-shot and one-shot NAS approaches, ZC-NAS is both~(i) time-efficient, eliminating the need for model training, and~(ii) interpretable, with proxy designs often theoretically grounded.~Despite rapid developments in the field, current SOTA ZC proxies are typically constrained to well-established convolutional search spaces.~
With the rise of Large Language Models shaping the future of deep learning, this work extends ZC proxy applicability to Vision Transformers (ViTs).~
We present a new benchmark using the Autoformer search space evaluated on 6 distinct tasks and propose Layer-Sample Wise Activation with Gradients information (L-SWAG), a novel, generalizable metric that characterizes both convolutional and transformer architectures across 14 tasks.~Additionally, previous works highlighted how different proxies contain complementary information, motivating the need for a ML model to identify useful combinations.~To further enhance ZC-NAS, we therefore introduce LIBRA-NAS (Low Information gain and Bias Re-Alignment), a method that strategically combines proxies to best represent a specific benchmark.~Integrated into the NAS search, LIBRA-NAS outperforms evolution and gradient-based NAS techniques by identifying an architecture with a 17.0\% test error on ImageNet1k in just 0.1 GPU days.\vspace{-0.5cm}
\end{abstract}

%% file: sec/1_intro.tex
\section{Introduction}
\label{sec:intro}

\gls{nas} optimizes neural networks for a given task and constraint replacing the costly trial and error design process~\cite{Zoph2017}.~Over the course of the years, it has gained attention for its ability to discover better performing and more efficient neural networks compared to hand-crafted ones~\cite{liu2018progressive, liu2018darts, elsken2019survey, pham2018enas, real2017evolution, xie2019snas, wu2019fbnet}.~With the advent of \gls{llm}s ruling the deep learning world with high accuracy, \gls{nas} is not seen anymore as a na\"ive tool for boosting performance.~It finds important applicability in real-world scenarios with hardware-aware models requiring pruning, different resource constraints, and memory footprint optimization~\cite{Li2024ZeroShot}. \\
Despite its advantages, the major drawback of NAS usually resides in the computationally demanding search process.~The first proposed multi-shot NAS methods involved training multiple candidate networks, requiring 
up to 28 days on 800 GPUs~\cite{Zoph2017}.~Subsequent one-shot approaches accelerated \gls{nas} by sharing candidate operations through a super-network (\cite{xu2019pcdarts, dong2019robust, zela2019robustifying, chen2019progressive, cai2019proxylessnas, cai2020onceforall}).~Weight-sharing (\cite{stamoulis2019singlepath, chu2021fairnas, guo2020singlepathoneshot, chen2020fasterseg}) further advanced by sharing also the parameters across different operations, improving memory efficiency.~Although the differentiable process reduced optimization time to a few GPU hours for tasks like Cifar-10, full training of the super-network is still required.~
Predictor-based methods remove the training of neural networks, avoiding the main drawbacks of heavy time and GPU resource consumption.~They achieve highly accurate performance estimation (\cite{liu2018progressive, luo2018neural}), but still require \emph{training the predictor}~\cite{wen2019neural, Dudziak2020BRP} over a NAS benchmark obtained through a costly data collection step constituted of thousands of networks trained until convergence. \\
Zero-shot~\gls{nas} methods therefore emerged with the promise of fully removing 
the data collection step by characterizing Deep Neural Networks (DNNs) through proxy metrics, an estimate of the performance of DNNs based on heuristic and theoretical results.~
This paper focuses on zero-shot NAS, which 
as pointed out in~\cite{Li2024ZeroShot}, brings two major advantages:~(i) time efficiency, as model training is eliminated, and (ii) interpretability, as the design of a proxy metric is usually inspired by some theoretical analysis of DNNs which helps in understanding the reason for their success.~
Since the first proposed metric~\cite{bhardwaj2021nnmass}, many proxies have been introduced in the literature. They usually characterize neural networks under three principles: (i) trainability (\cite{Li2023Zico, lee2019snip, zhang2022gradsign}), (ii) generalization~\cite{chizat2019lazy, croze}, and (iii) expressivity (\cite{lin2021zen, peng2024swapnas, mellor2021neural, bhardwaj2021nnmass}). 
Most recent works often propose new metrics grounded in either theoretical frameworks~\cite{chen2021nas, lga, bhardwaj2021nnmass} or heuristic approaches~\cite{mellor2021neural, zhou2024hytas}.~This frequently results in a large variety of metrics that leave unclear the reasons for their effectiveness.~Moreover, despite a few efforts~\cite{nasbenchsuitezero}, these proxies are often evaluated on different setups, hindering their true contribution and relations with respect to the state-of-the-art.~Evaluation is typically performed on a few search spaces (\eg, NAS Bench201~\cite{dong2020nasbench201}), which provides limited insight since most metrics show strong correlation results within these spaces.

Therefore, different from other studies, 
we first of all test all existing metrics under the same setup and include in our analysis the ViT search space.~Our first goal is to expand the scope of applicability of proxy metrics, 
opening the road to nowadays topics, like video understanding, which could be addressed with ViTs.~
Our experiments reveal that in the ViT search space, many ZC-proxies struggle to outperform basic metrics like \# Parameters.~In response, we introduce the Layer Sample-Wise Activation with Gradients (L-SWAG) metric, which not only surpasses \# Parameters on the ViT search space but also outperforms existing metrics across several benchmarks, including the challenging TransNasBench~\cite{tnb101}, where most other metrics fall short.
To properly handle the different characteristics of search-spaces we developed Low Information Gain and Bias Re-alignment (LIBRA)-NAS, a novel ensemble algorithm. Observations indicate that certain search spaces may favor gradient-based metrics, while others are better suited to gradient-free ones.~Some metrics tend to introduce a strong bias toward cell size, while others penalize networks that converge quickly.~
Additionally, different proxy metrics often contain complementary information highly dependent on the chosen benchmark~\cite{nasbenchsuitezero}.~This phenomenon motivates the need for a ML model that can identify effective combinations of proxy metrics based on the specific requirements of each benchmark.
To summarize, our contributions are:
\begin{itemize}
    \item We train and evaluate 2000 ViT architectures on six different tasks, and evaluate all existing ZC-proxy metrics on this new benchmark, adapting metrics formulated only for ReLU networks also to GeLU ones.
    \item We present L-SWAG metric, which captures a layer-wise trainability and expressivity of DNNs and positively correlates on the ViT search space, improving state-of-the-art Spearman $\rho$ correlation on several benchmarks.
    \item We propose LIBRA, a new ensemble algorithm to be used when exceptionally high correlation, not currently attainable by a single proxy, is needed.~LIBRA   
    combines metrics based on complementary proxy information and on benchmark biases.~In the NAS search, LIBRA beats previous RL and evolution methods finding an architecture with 17.0 \% test error on ImageNet1k in 0.1 GPUdays. 
\end{itemize}


%% file: sec/2_related_works.tex
\section{Related works}
\label{sec:related_works}

\begin{figure*}[t]
  \centering
     \includegraphics[width=17cm, height=5.3cm]{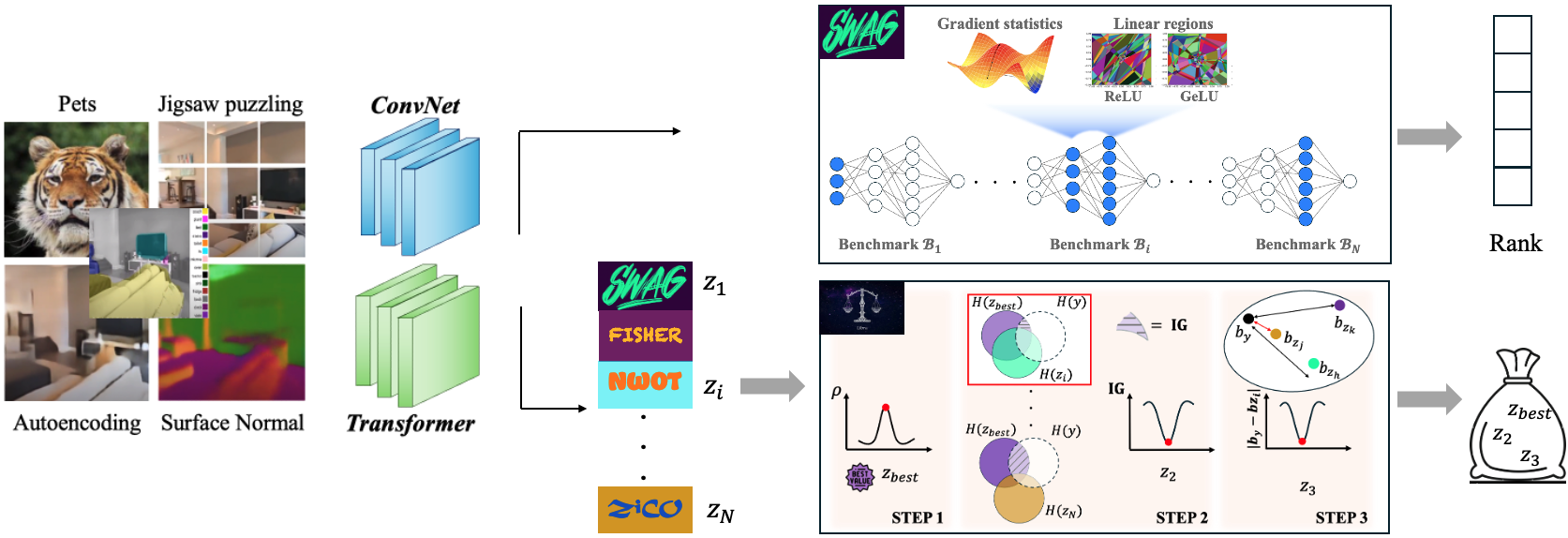}
    \caption{Our approach applies to different task types of architectures. L-SWAG takes as input a batch of images and a DNN, extracts the gradient statistics, and counts the \# of linear regions in a layer-wise fashion. The relevant layers are identified una-tantum, before running the metric and are specific for each benchmark. L-SWAG outputs a rank of the architectures. LIBRA takes as input the pre-computed ZC-proxy metrics for a given benchmark. It has three steps: (i) selects the best performing one according to their correlation $\rho$. (ii) Computes the information we gain over the validation accuracy $y$ given $z_{best}$ and each other $z_i$, and selects the $z$ leading to the lowest validation accuracy. (iii) Select $z_3$ with the closest bias to $y$. LIBRA outputs the 3 identified metrics.}
    \label{fig:Method}\vspace{-0.5cm}
\end{figure*}

Zero-shot NAS designs proxies that can rank architectures' accuracy given the network at the initialization.~They require only a single forward pass through the network, taking a few seconds~\cite{mellor2021neural}, and do not involve parameters update nor gradient descent.~Existing works usually focus on proxies related to~(i) expressivity, reflected by the number of linear regions over the input space in ReLU networks (\cref{sec:grad_free}),~ii) generalization and,~(iii) trainability through gradient properties (\cref{sec:grad_based}).
Recent works address a deeper understanding of existing proxies and propose new aggregation methods to get a more comprehensive characterization of DNNs through proxy combination (\cref{sec:metric_aggregation}).
\subsection{Gradient based proxies}\label{sec:grad_based}
Inspired by pruning-at-initialization techniques,~\cite{abdelfattah2021zerocost} formulates a proxy that estimates each weight parameter's importance by analyzing its gradient.~GradSign~\cite{zhang2022gradsign} analyzes the sample-wise optimization landscape and defines a proxy for the upper bound of the loss.~Fisher~\cite{theis2018faster} uses approximated second-order gradients (\ie empirical Fisher Information Matrix EFIM) at a random initialization point.~Although it correlates well on certain search spaces where other measures fail (\eg Tnb101-micro\_AE), the EFIM is a valid approximation only if the model's parameters are a Maximum Likelihood Estimation, an invalid assumption at a random initialization point, as highlighted in~\cite{zhang2022gradsign}.~SNIP~\cite{lee2019snip} integrates the values of the parameters to gradients properties, GraSP~\cite{wang2020picking} considers both the first order and the second order derivatives of the gradients, while JacobCov~\cite{lopes2021epenes} leverages gradients over the input data instead of parameters.~GSNR~\cite{sun2023unleashing} proposes a proxy based on the gradient Signal to Noise Ratio (SNR) theoretically proved to be linked to generalization and convergence.~ZiCO~\cite{Li2023Zico} characterizes network trainability, convergence, and generalization through the mean and the standard deviation of gradients.~
Our L-SWAG measure is strictly related to~\cite{Li2023Zico}, but differently from ZiCO, we (i) discard the mean of gradients through theoretical (\cref{sec:swag}) and empirical motivations (\cref{tab:ablation_swag}) 
and (ii) provide a layer-wise formulation, showing (\cref{fig:tnb101_ae_layers}) how specific layers statistics are more informative than others.~Finally, (iii) our metric does not fail on the ViT search space.~As shown in~\cref{fig:corr_mat1000}, we attribute the success to the inclusion in SWAG of an expressivity term. 

\subsection{Gradient-free proxies}\label{sec:grad_free}
Gradient-free proxies entirely remove backward propagation and focus on the expressivity or topology properties of DNNs represented as graphs.~\cite{mellor2021neural, peng2024swapnas} study the number of linear regions after ReLU activations.~NWOT~\cite{mellor2021neural} computes the Hamming distance between binary codes (rows in a standard activation pattern) obtained from ReLU patterns and defines a metric ``distinctive for DNNs that perform well".~
Despite the empirical proof of correlation, 
NWOT struggles in search-spaces with lower accuracies.~
Zen-Score~\cite{lin2021zen} is an almost ZC proxy metric.~It measures expressivity through a few forward inferences on randomly initialized networks using random Gaussian inputs.~
As highlighted in~\cite{lin2021zen}, it is not mathematically defined on irregular search spaces as DARTS~\cite{liu2018darts} and Randwire~\cite{randwire}.~Finally, NAS-Graph~\cite{huang2024graph} converts DNNs into graphs and uses the average degree of nodes as a proxy.\vspace{-0.2cm}

\subsection{Metric aggregation methods}\label{sec:metric_aggregation}

NAS-Bench-SuiteZero~\cite{nasbenchsuitezero} evaluates for the first time many proxy metrics under a great variety of tasks through fair conditions and a unified codebase. We extend this effort by including recently proposed metrics (\cite{Li2023Zico, peng2024swapnas, lee2024aznas}) and a ViT search space over six different tasks. 
Nas-Bench-SuiteZero uses correlation analysis and information theory to identify complementary information and biases in each proxy. 
Differently, we propose a way of integrating metrics that does not involve a predictor (that cannot be considered zero-shot) and formulate a ``bias matching" technique which we empirically show improves over the authors ``bias mitigation". 
Te-NAS~\cite{chen2021nas} uses both the number of linear regions~\cite{hanin2019complexity, xiong2020number} and the condition number of Neural Tangent Kernel (NTK)~\cite{jacot2018neural, lee2019wide}. However, not only calculating NTK is computationally demanding~\cite{ning2021evaluating}, but a recent work~\cite{lga} proves how the hypothesis of NTK theory 
does not apply to modern DNNs.~Therefore no foundations are available on why NTK at initialization should be used.~Moreover, Te-NAS exploits the \# of linear regions on what~\cite{peng2024swapnas} calls a ``standard activation pattern" which has proven to fail on input of large dimensions.~
T-CET~\cite{tcet} revisits existing metrics providing new theoretical insights to formulate a new proxy comprising compressibility, orthogonality and topology of neural networks.~
They integrate a layer-wise NWOT formulation into the SNR, offering a new interpretation of ZiCO’s $\sigma$ component from a compressibility perspective. This approach helps explain why ZiCO’s theoretical foundations, developed for linear networks, hold for more complex nonlinear networks but does not address the need for ZiCO's $\mu$ component. 
In this study, we show why $\mu$ should be discarded, giving theoretical and empirical proof.~Differently from T-CET, we provide a clear heuristic to select the needed layers for $\sigma$ computations.~
AZ-NAS~\cite{lee2024aznas} advocates for using an ensemble of proxies instead of a single one and introduces 
four proxies tackling: expressivity, trainability, progressivity, and complexity.~
In AZ-NAS a ViT search space is included in the experiments.~However, the evaluation is done by integrating the proxy directly into the NAS search, which, in our view, does not adequately assess the effectiveness of the proxies. The ViT search space~\cite{autoformer} is known to yield well-performing subnetworks, all achieving between the best accuracy and within 2\% of the best accuracy. As a result, the ability of metrics to guide the search is difficult to evaluate, as random search also yields strong performance (\cf supp. material). In contrast, we also conduct a correlation analysis with the validation accuracies obtained by training 2,000 networks on each task.\vspace{-0.3cm}

%% file: sec/3_methods.tex
\section{Method}\label{sec:method}
In this section, we describe the overall framework depicted in~\cref{fig:Method}.~
Our first goal is to efficiently rank architectures on a ViT search-space, keeping strong performance and good generalization on commonly deployed search spaces.~To achieve this we formulate L-SWAG, capturing trainability and expressivity for ReLU and GeLU networks (\cref{sec:swag}).~We present its key components and show the benefits of a layer-wise formulation.~Our second goal is to design a ML model to properly combine existing metrics depending on the characteristics of the considered benchmark.~To this aim, we introduce LIBRA-NAS (\cref{sec:libra}), which analyses complementary information and biases. 

\subsection{L-SWAG-Score}\label{sec:swag}
The design of our metric is motivated by three main findings mapped in the blue components:
{{\fontsize{8pt}{10pt}\selectfont\vspace{-.1in}
\begin{equation}
    \text{L-SWAG} = \overbrace{\sum_{l=\hat{l}}^{\textcolor{blue}{\hat{L}}} \log \left( \sum_{w \in \mathbf{\theta}_l} \frac{\textcolor{blue}{1}}{\sqrt{\operatorname{Var}(|\nabla_w \mathcal{L}(\mathbf{X_i}, \mathbf{y_i}; \Theta)|)}} \right)}^{\substack{\textcolor{gray}{\Lambda^{\hat{L}}}}} 
    \times  \overbrace{\left| \textcolor{blue}{\hat{\mathbb{A}}_{\mathcal{N}, \theta}^{\hat{L}}} \right|}^{\substack{\textcolor{gray}{\Psi_{\mathcal{N}, \theta}^{\hat{L}}}}}
    \label{eq:L-SWAG}
\end{equation}
}
where $\Theta$ denotes the initial parameters, $\theta_l$ the parameters of the $l^{th}$ layer, $w$ represents each element in $\theta_l$, $\hat{L}$ an intermediate layer in the network with maximum depth $L$, $\mathbf{X_i}, \mathbf{y}_i$ the input batch and corresponding labels from the training set, and $\Psi_{\mathcal{N}, \theta}^{\hat{L}}$ the component defined in~\cref{def:l-swap}. 
The first finding is related to the formulation of $\Lambda$ in~\cref{eq:L-SWAG} and to the presence of 1 instead of $\mu$ proposed by~\cite{Li2023Zico} at the numerator.
We first analyzed ZiCO, which in essence, advocates for choosing a candidate that maximizes the expected gradient in each of its layers, while keeping variance low. This choice is motivated in~\cite{Li2023Zico} by Theorem~3.1, which proves a bound on the empirical error of a linear regressor. We argue that, while the latter principle is correct (further motivated by Theorem~3.3 and~3.5 in~\cite{Li2023Zico}), the former is not. 
Given a training set $\mathbb{S}$ with $M$ samples:\vspace{-0.2cm}
    \begin{align}
    \mathbb{S} = \left\{ ({\vx}_i, y_i) \mid i = 1, \dots, M, \; \vx_i \in \mathbb{R}^d, \; y_i \in \mathbb{R}, \; \right. \notag \\
    \left. \|{\vx}_i\| = 1, \; |y_i| \leq R, \; M > 1 \right\}
    \end{align} 
    with $R > 0$ and $||\cdot||$ denoting the L2-norm of a given vector, ${\vx_i}\in \mathbb{R}^d$ the $i^{th}$ input samples normalized by its L2-norm, and $y_i$ the corresponding label. 
    Let's define a linear model $f = {\va^T \vx}$ optimized with an MSE-based loss function $\mathcal{L}$:\vspace{-0.3cm}
    \begin{align}
        \min_\va\sum_i\mathcal{L}(y_i, f(\vx_i; \va)) = \min_\va\sum_i\frac{1}{2}(\va^T\vx_i - y_i)^2
    \end{align}
    where $\va\in\mathbb{R}^d$ is the initial weight vector of $f$.
    Let's denote with $g(\vx_i)$ the gradient of $\mathcal{L}$ w.r.t $\va$, and as $g_{j}(\vx_i)$ the j-th element of $g(\vx_i)$. The mean value $\mu_j$ and standard deviation $\sigma_j$ of $g(\vx_i)$ are obtained as follows:
    \begin{align}
        \mu_j = \frac{1}{M}\sum_i^Mg_j(\mathbf{x_i}) \quad \sigma_j = \sqrt{\frac{1}{M}\sum_i^M(g_j(\mathbf{x_i})-\mu_j)^2}
    \end{align}
\begin{theorem}\label{theorem:3.1}
    Given the linear regressor $f(\va, \vx)$ with trainable parameters $\va = (a_j)_{j = 1}^M$, let $g(\vx_i) = (g_j(\vx_i))_{j = 1}^d$ be the gradient of $\va$ w.r.t. to $\vx_i$, and $\hat{\va} = \va -\eta\sum_i g_j(\vx_i)$ the updated parameters with learning rate $\eta$. Denote $\mu_j = \frac{1}{M}\sum_i g_j(\vx_i)$, $\sigma_j = \sqrt{\sum_i (g_j(\vx_i) - \mu_j)^2}$. Then, for any $\eta$, the total training loss $\mathcal{L}_f(\mX, \vy; \hat{\va}) = \frac{1}{2}\sum_i (\hat{\va}^\top \vx_i - y_i)^2$ of $f$ is bounded by:\vspace{-0.2cm}
    \begin{equation}
         \mathcal{L}_f(\mX, \vy; \hat{\va}) \le \frac{1}{2}\left(M\sum_{j = 1}^d \left[\sigma_j^2 + ((M\eta - 1)\mu_j)^2\right]\right).
    \end{equation}\vspace{-0.6cm}
\end{theorem}
\begin{proof}
    \cf supplementary material.
\end{proof}




No other theorems in~\cite{Li2023Zico} support the need for $\mu$ for nonlinear networks, and as we show in~\cref{tab:ablation_swag} (and with an empirical validation of Th.~1 in supp. material) our formulation in \cref{eq:L-SWAG} with 1 instead of $\mu$ benefits performance.\\
\textbf{Layer contribution.   }
Our summation in~\cref{eq:L-SWAG} starts with $\hat{l}$ and ends with $\hat{L}$, two intermediate layers in the network. This differs from usual formulations~\cite{Li2023Zico, zhou2024hytas, zhang2022gradsign}
which usually treats equally the statistics of all layers in a network. However, previous studies already highlighted how not all layers bring equal contributions in terms of gradient statistics. 
In~\cite{raghu2017ontheexpressive}, the authors emphasize that ``trained DNNs are more sensitive to weights in the lower (initial) layers.". In~\cite{bjorck2018UnderstandingBN} several experiments show a larger standard deviation of gradient for lower layers. In~\cite{zhou2024hytas} the authors highlight how ZiCO has a  ``heavy reliance on the \# of layers".
All these hints motivated us in analyzing the statistics of the gradients layer-wise, to answer the following question: \emph{Can we remove some layers from the statistic extraction? Are all layers of equal importance?} 
Our approach simply consists of plotting the statistics of the gradients for 1000 randomly sampled DNNs at initialization.~\cref{fig:tnb101_ae_layers} reflects what is the mean intensity and standard deviation of the $\sigma_j$ of the gradient through percentiles, where a percentile is obtained following this rule:\vspace{-0.3cm}
{\small
\begin{align}
    \texttt{perc = int}\left(\frac{\texttt{l}}{\texttt{D}} \texttt{*100} \, \texttt{//} \, \left(\frac{\texttt{100}}{\texttt{PERC\_BINS}}\right)\right)
\end{align}}
with $l=1,\dots,L$, and \texttt{\small{PERC\_BINS = 10}}, to properly average results of DNNs with different depths. We also checked the influence of depth by clustering networks based on $L$, as the influence of $\sigma_j$ may vary, but we did not find significantly different behaviors (\cf S.M.). All benchmarks share the same behavior and report spikes on specific percentiles (see~\cref{fig:tnb101_ae_layers}, and S.M for all benchmark results). We found that by considering as $\hat{l}$ and $\hat{L}$ the beginning and the end of spikes respectively, a huge improvement in terms of rank correlation is experienced. This can be visualized in~\cref{fig:evo_rank}, where selecting only specific percentiles, large improvements, depicted by yellow regions, in the rank correlation are experienced. This layer-wise selection moreover speeds-up the metric calculation (see~\cref{tab:ablation_swag}).\vspace{-0.5cm}
\begin{figure}[ht]
    \centering
    \begin{subfigure}[b]{0.5\textwidth}
        \centering
        \includegraphics[width=0.8\textwidth]{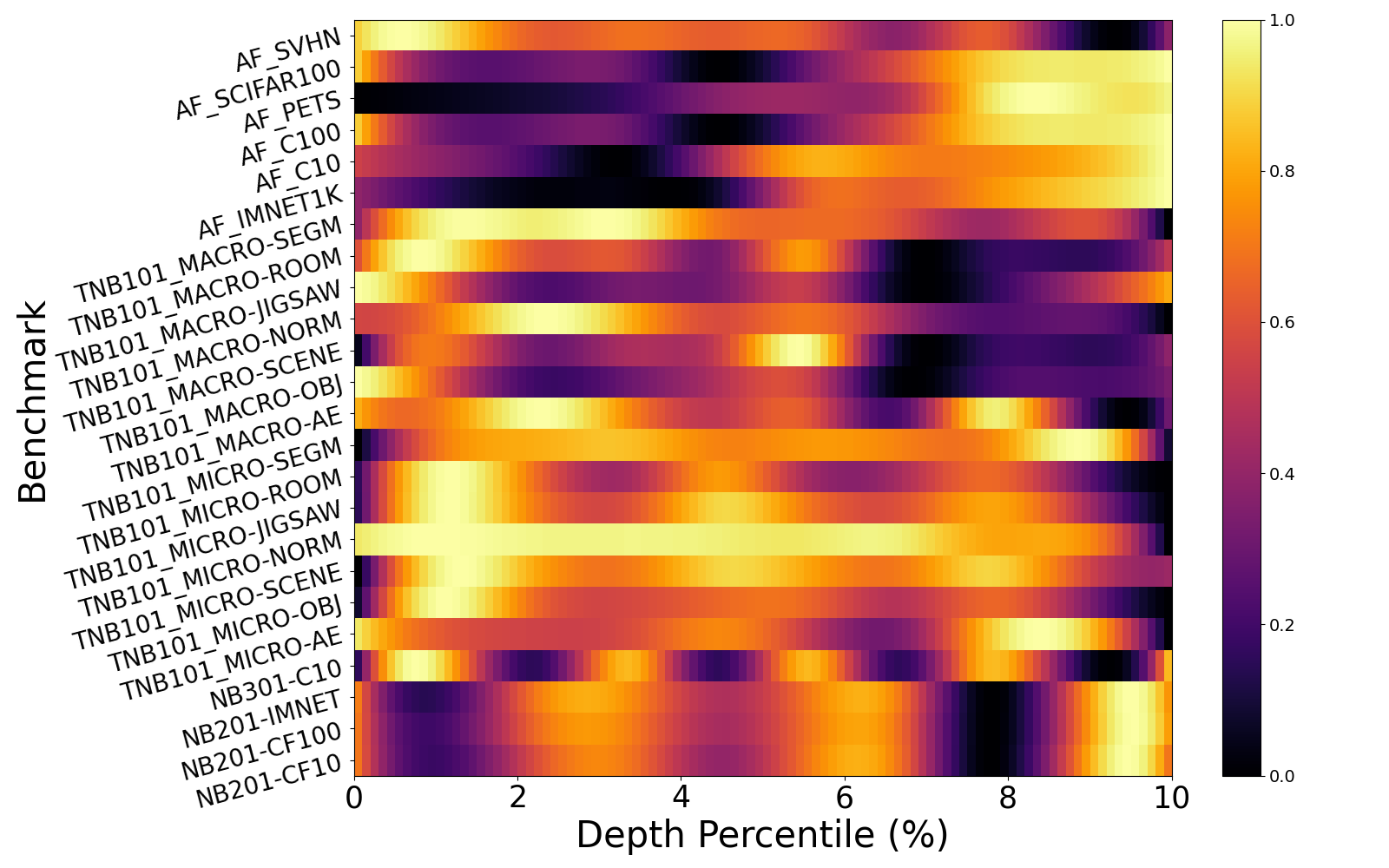}
        \caption{Normalized Spearman $\rho$ correlation for each selected percentile. Larger values occur in correspondence to the spikes of the $\sigma$ statistics of the gradient in the below graph.}\vspace{-0.4cm}
        \label{fig:evo_rank}
    \end{subfigure}
    
    \vspace{0.5cm} 
    
    \begin{subfigure}[b]{0.5\textwidth}
        \centering
        \includegraphics[width=5.5cm, height=4cm]{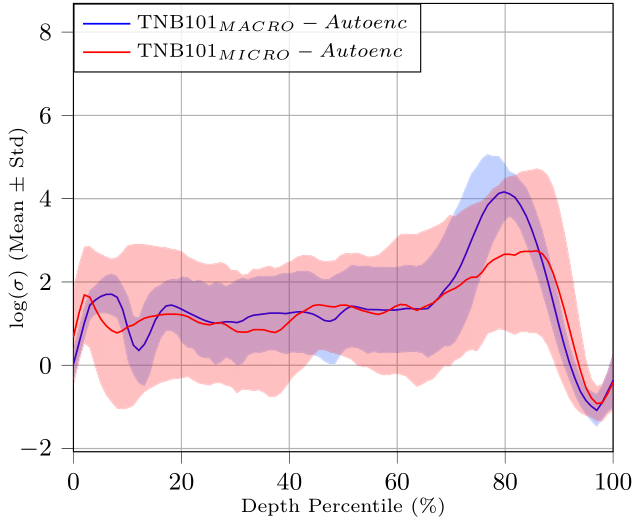}
        \caption{Average gradient statistics across 1000 networks sampled from the Autoencoder Micro/Macro search spaces.}\vspace{-0.2cm}
        \label{fig:tnb101_ae_layers}
    \end{subfigure}
    \caption{Empirical motivation for our layer selection strategy.
    }\vspace{-0.6cm}
    \label{fig:stacked_subfigures}
\end{figure}

        
        
    

\paragraph{Expressivity.}
Inspired by~\cite{peng2024swapnas, mellor2021neural}, we assess the expressivity of DNNs over a batch of input samples. To this aim, we deploy the cardinality of activation patterns of ReLU and, for the first time, of GeLU networks on a layer-wise partition.\vspace{-0.1cm} 

\begin{definition}
\textbf{(Sample-Wise Activation Patterns).} Given a ReLU or GeLU deep neural network \(\mathcal{N}\), a set \(\theta\) of randomly initialized parameters, and a batch of inputs with \(S\) samples, the set of layer-wise sample-wise activation patterns \(\hat{\mathbb{A}}_{\mathcal{N}, \theta}^{\hat{L}}\) is defined as follows:
\begin{align}
    \hat{\mathbb{A}}_{\mathcal{N}, \theta}^{\hat{L}} = \left\{ \vp^{(l)} : \vp^{(l)} = \mathds{1}(p_s^{(l)})_{s=1}^S, \, l \in \{1, \dots, \hat{L}\} \right\}
\end{align}
where $p_s^{(l)}$ denotes a single-post activation value from the $s^{th}$ sample at the $l^{th}$  intermediate value. $\hat{L} \in {1, \dots, L}$ with $L$ layers in the network. 
\end{definition}
$\mathds{1}(p_s^{(v)})_{s=1}^S$ is a vector containing binarised post-activation values across all samples in $S$.
We can now define the layer-wise SWAP-Score:\vspace{-0.2cm}
\begin{definition}
    Given a layer-wise SWAP set \(\hat{\mathbb{A}}_{\mathcal{N}, \theta}^{\hat{L}}\), the layer-wise SWAP-Score \(\Psi\) of a network \(\mathcal{N}\) with a set \(\theta\) of randomly initialized parameters is defined as the cardinality of the set:
    \begin{equation}
        \Psi_{\mathcal{N}, \theta}^{\hat{L}} = \left| \hat{\mathbb{A}}_{\mathcal{N}, \theta}^{\hat{L}} \right|
    \end{equation}
    \label{def:l-swap}
    \vspace{-0.3cm}
\end{definition}
On a practical basis, layer-wise SWAP represents the ``practical expressivity" of each layer. 
To summarize, L-SWAG combines through multiplication a layer-wise trainability measure $\Lambda^{\hat{L}}$ and an expressivity measure $\Psi_{\mathcal{N}, \theta}^{\hat{L}}$. The reason for multiplying and not adding them is deeply motivated in~\cite{tcet} and summarize in our supp. material. As we will show in~\cref{sec:exp}, both components are needed to perform well on standard benchmarks, and on ViT search space. \vspace{-0.2cm}

\subsection{LIBRA-NAS}\label{sec:libra}
This section introduces our \textbf{L}ow \textbf{I}nformation gain and \textbf{B}ias \textbf{R}e-\textbf{A}lignment (LIBRA) (\cref{algo:LIBRA}) for NAS which we deploy to merge different proxies.\vspace{-0.1cm}
\begin{algorithm}
    \SetAlgoNlRelativeSize{0}
    \caption{LIBRA}\label{algo:LIBRA}
    \textbf{Input:} Set of proxies $Z$ with their correlation $\rho$ over benchmark $\mathcal{B}_{ij} = (\text{search space } \mathcal{S}_i, \text{dataset } \mathcal{D}_j)$, and $b_{val}$, the bias of the validation accuracy on $\mathcal{B}_{ij}$.\\
    \textbf{Output:} Subset $Z_{\text{libra}}$ for $\mathcal{B}_{ij}$.
    \BlankLine

    \For{each $\mathcal{B}_{ij}$}{
        \textnormal{Select the proxy $z_k$ with the highest correlation $\rho_{best}$}\;
        \textnormal{Initialize empty lists, $IG\_list$ and $B\_list$}\;
        
        \For{$z_h \in \{Z \setminus z_k\}$}{    
            \If{$\rho_{best} - 0.1 < \rho_h \leq \rho_{best}$}{
                \textnormal{Compute Information Gain IG($z_h$) according to~\cref{eq:info_gain}}\;
                Add IG($z_h$) to $IG\_list$\;
            }
        }
        $z_1 \gets z_k$\;
        $z_2 \gets \arg \min IG\_list$\;
        \For{$z_h \in \{Z \setminus \{z_1, z_2\}\}$}{
            \If{$\rho_{best} - 0.1 < \rho_h \leq \rho_{best}$}{
                \textnormal{Compute the bias $b_{z_h}$ for $z_h$}\;
                Add $|b_{val} - b_{z_h}|$ to $B\_list$\;
            }
        }
        $z_3 \gets \arg \min B\_list$\;
    }\vspace{-0.0cm}
\end{algorithm}\vspace{-0.3cm}
Given a set of pre-computed proxies $Z$ and pre-computed bias values $b$, easily obtainable thanks to works like~\cite{nasbenchsuitezero} and ours, LIBRA-NAS outputs three proxies which are useful combinations to boost the performance on a given benchmark $\mathcal{B}_{ij}$. The bias, although in principle could be of any kind \eg \# of convolutional layers, \# of skip connection, etc., in our implementation is represented by the \# of parameters. Each bias value is computed by checking the Pearson correlation between the rank induced by the validation accuracy/proxy metric considered, and the rank induced by the bias (\cf supp material for values).~Following the entropy and information gain definition provided~\cite{nasbenchsuitezero}, given a search space $\mathcal{S}$, let $\mathcal{Y}$ be the uniform distribution of validation accuracies over $\mathcal{S}$, and $y$ be a random sample from $\mathcal{Y}$.~Now let $\mathcal{Z}$ be the uniform distribution for the proxies and $z$ a sample from it.~Given the entropy function $H(\cdot)$ the information gain between two proxies is obtained with: \vspace{-0.2cm}
\begin{align}
    \textbf{IG}(z_{j}) &= H(y | z_i) - H(y | z_i, z_j). 
     \label{eq:info_gain}
\end{align}
The proposed algorithm selects the best proxy metric for the given $\mathcal{B}_{ij}$.~Subsequently, among those performing in the specified range (0.1 in our case, empirically selected) it computes $\textbf{IG}(z_{j})$ and selects the one leading to the lowest information gain.~Intuitively, $\textbf{IG}$ represents the additional information gained about $y$ when $z_j$ is disclosed, given that the values of $z_i$ are already known.~While the motivation for minimizing this value is largely heuristic, we suggest that minimizing (rather than maximizing) it yields optimal results.~This approach can be thought of as analogous to ``overfitting", as we are selecting metrics that capture the same aspects of the search space.~Then, the third metric is chosen among the top-performing ones sharing a similar bias the validation accuracy has.~Other approaches mitigate the bias by removing it~\cite{nasbenchsuitezero}.~We rather show with ablations~\cref{tab:ablation_LIBRA} it gives the best performance indulging the same bias the metric we are estimating has. \vspace{-0.3cm}

%% file: sec/4_experiments.tex
\section{Experiments}\label{sec:exp}\vspace{-0.2cm}
We conduct the following experiments:~(i) evaluation of Spearman $\rho$ correlation of L-SWAG on multiple NAS benchmarks, including the ViT search space~\ref{sec:exp_lswag},~(ii) evaluation of LIBRA-NAS $\rho$ on state-of-the-art benchmarks and comparison with other proxy-merging methods,~(iii) illustration of L-SWAG-based and LIBRA-based zero-shot NAS on Cifar-10 and ImageNet~\cref{sec:exp_libra},~(iv) ablations of each component for both contributions~\cref{sec:ablation}.\\
\textbf{Experimental Settings.}~We compare L-SWAG with all metrics considered in~\cite{nasbenchsuitezero} and with recent SOTA approaches ZiCO, SWAP and reg\_SWAP. LIBRA is evaluated against all existing, to the best of our knowledge, types of zero-shot merging techniques. Our codebase is based on NASBench-SuiteZero, and all experiments were run on a single RTX 3090Ti. The gradient statistics extraction takes 31 mins for 1000 ViTs with \# params.~$\in$15-35M, on ImageNet with 224$\times$224 resolution. The memory occupation is$\sim$10 GB. After selecting the layers, the L-SWAG calculation takes $\sim$4 minutes.
All main results are obtained on 1000 architectures using a batch of 64 for all benchmarks but TransBench-101, which for high memory usage required a batch of 32. Results for the whole search-space can be found in the supp. material.\\ 
\textbf{Datasets. }
We evaluate our proxies across different tasks: NASBench-201 (Cifar-10, Cifar-100 and ImageNet16-120), NASBench-101~\cite{nasbench101} (Cifar-10), NASBench-301~\cite{nasbench301} (Cifar-10), TransNAS-Bench-101 Micro and Macro~\cite{tnb101} (Jijsaw, Object and Scene Classification, Autoencoder, Room Layout, Surface Normal, Semantic Segmentation).~We chose these benchmarks following~\cite{nasbenchsuitezero}.~We re-produced all results as many works~\cite{Li2023Zico, peng2024swapnas, tcet, lee2024aznas, chen2021nas} did not run experiments on TransBench-101, NasBench-301 and NasBench-101.~L-SWAG and all metrics are then also evaluated 2000 multiple times trained networks sampled from the Autoformer~\cite{autoformer} Small search-space. These networks were trained on: ImageNet, Cifar10, Cifar100, Pets, SVHN, and Spherical-Cifar100. We included a ViT search space to expand the scope of applicability of proxy metrics (\cf supp. material for details on the training procedure for ViT architectures and full description of datasets). 
\subsection{L-SWAG Ranking Consistency}\label{sec:exp_lswag}

\begin{figure*}[t]
  \centering
    \includegraphics[width=16cm, height=6cm]{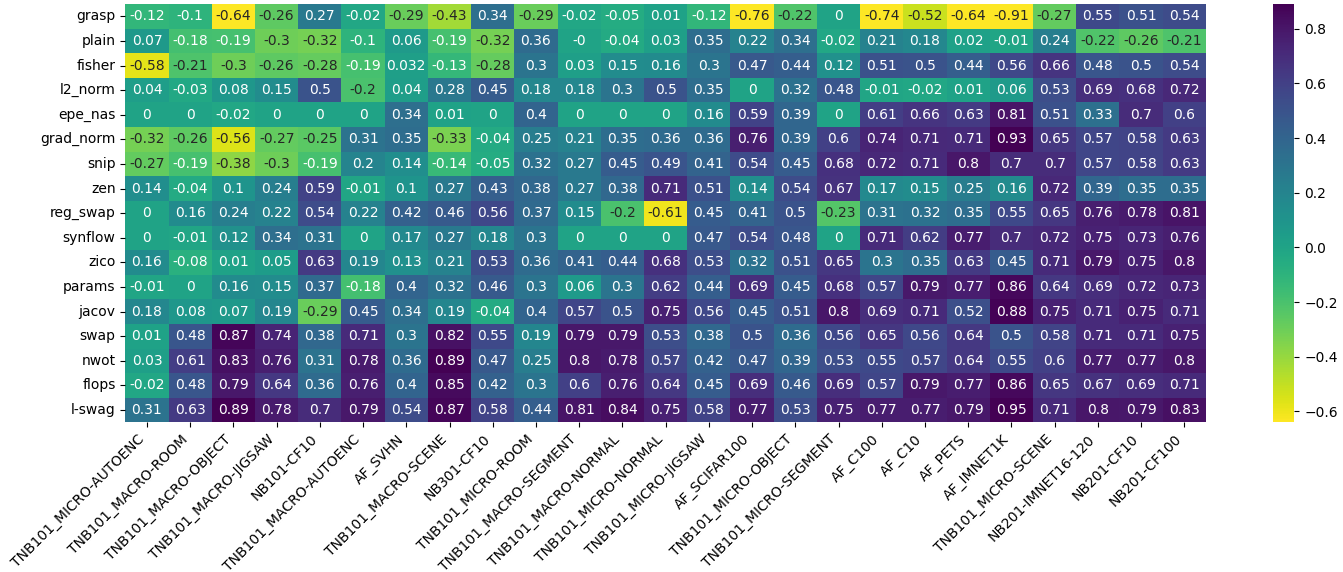}
    \caption{Spearman rank correlation coefficient between ZC proxy values and validation accuracies. Results were obtained from 5 multiple runs. Rows and columns are ordered based on the mean scores.}
    \label{fig:corr_mat1000}\vspace{-0.3cm}
\end{figure*}

\begin{figure}[t]
  \centering
    \includegraphics[width=8.cm, height=4.5cm]{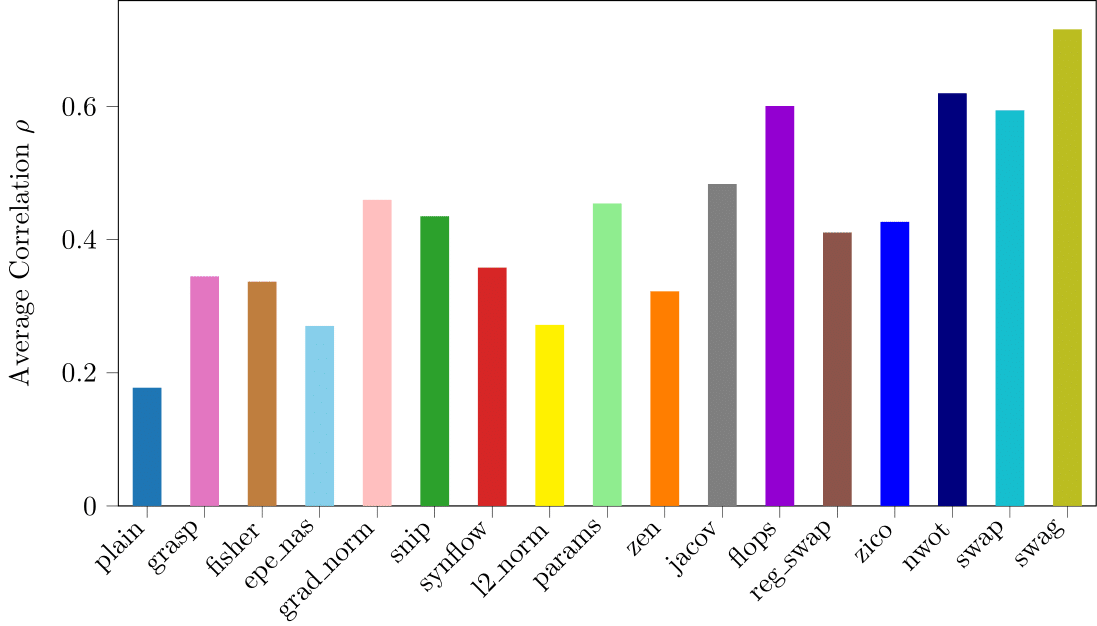}
    \caption{Average Spearman $\rho$ coefficient of ZC proxies across different search spaces.}\vspace{-0.8cm}
    \label{fig:avg_bar}
\end{figure}

We show in~\cref{fig:corr_mat1000} and~\cref{fig:avg_bar} a quantitative comparison between L-SWAG and state-of-the-art ZC proxies.~\cref{fig:corr_mat1000} details the Spearman's $\rho$ correlation  over every benchmark, while~\cref{fig:avg_bar} highlights the average performance across benchmarks, proving the better performance consistency of L-SWAG with an average correlation of $\rho_{l-swag}=0.72$ over the second best $\rho_{nwot} = 0.62$. All values were obtained selecting specific percentiles based on the principle illustrated in~\cref{sec:swag}. We can see that L-SWAG achieves the best ranking consistency across several benchmarks, outperforming others by a large margin. In particular, we improve over tnb101 Macro jigsaw/normal, on nb101, nb301, on tnb101 Micro room/jigsaw. We also noticed however, that despite improving by a fair margin with respect to most ZC proxies on tnb101\_micro autoencoder, our result still underperforms fisher in this complex task. Focusing on competitors strictly related to our measure, \ie ZiCo and SWAP, a difference is experienced particularly on tnb101 Macro object/room/jigsaw, where ZiCO does not correlate, and on tnb101 Micro (for all tasks), where SWAP's $\rho$ diminishes. 
In comparison to the second-best metric, NWOT (excluding FLOPs), we observe that NWOT’s performance drops significantly when shifting from a Macro to a Micro search space, whereas this drop is much less pronounced with L-SWAG. A similar trend is observed with SWAP, which is not surprising given the close relationship between these metrics. We suggest that NWOT's decline in performance is due to its reliance solely on data separability and the assumption that this characteristic correlates with “well-performing networks.” 
Within the Autoformer search space, L-SWAG is the only metric that consistently outperforms or matches the performance of the competitive, simple proxy of parameter/FLOP count. It also shows improvement over the more commonly used NB201 search space, though we consider this search space less informative, as most metrics perform well in it. When integrated into the NAS framework (\cref{tab:libra_swag_nas}), L-SWAG identifies better architectures than its competitors at significantly lower costs, regardless of the specific task or search space. This demonstrates the method’s adaptability across diverse network architectures.


\begin{table*}
\centering
\scriptsize
\renewcommand{\arraystretch}{1.2} 
\setlength{\tabcolsep}{4pt}
\begin{tabular}{> {\arraybackslash}m{1.5cm} >{\centering\arraybackslash}m{0.4cm} >{\centering\arraybackslash}m{0.4cm} >{\centering\arraybackslash}m{1.1cm}  >{\centering\arraybackslash}m{0.6cm} > {\centering\arraybackslash}m{0.6cm}  > {\centering\arraybackslash}m{0.7cm} >{\centering\arraybackslash}m{0.5cm} >{\centering\arraybackslash}m{0.4cm} >{\centering\arraybackslash}m{0.5cm} >{\centering\arraybackslash}m{0.4cm} >{\centering\arraybackslash}m{0.5cm} >{\centering\arraybackslash}m{0.5cm}  > {\centering\arraybackslash}m{0.7cm} >{\centering\arraybackslash}m{0.5cm} >{\centering\arraybackslash}m{0.4cm} >{\centering\arraybackslash}m{0.5cm} >{\centering\arraybackslash}m{0.4cm} >{\centering\arraybackslash}m{0.5cm} >{\centering\arraybackslash}m{0.5cm}}
\hline
\multicolumn{1}{c}{} & \multicolumn{3}{c}{{\scriptsize{NB201}}} & \multicolumn{1}{c}{\scriptsize{NB101}} &\multicolumn{1}{c}{\scriptsize{NB301}} & \multicolumn{7}{c}{\scriptsize{TNB101-Micro}} & \multicolumn{7}{c}{\scriptsize{TNB101-Macro}} \\
\cmidrule(lr){2-4} \cmidrule(lr){5-5} \cmidrule(lr){6-6} \cmidrule(lr){7-13} \cmidrule(lr){14-20}
\multirow{1}{*}{} & C10 & C100 & IN16-120& C10 &C10& AE & Room & Obj. & Scene & Jig. & Norm. & Segm. & AE & Room & Obj. & Scene & Jig. & Norm. & Segm.\\ 
\hline\hline

TE-NAS & 0.70 & 0.67 & 0.64 & 0.12 & 0.37 & -0.41 & 0.51 & 0.37 & 0.25 & 0.13 & 0.10 & 0.34 & -0.55 & 0.05 & 0.13 & 0.28 & 0.65 & 0.61 & 0.03 \\ 
T-CET & 0.77 & 0.80 & 0.81 & 0.23 & 0.42& 0.31 & 0.34 & 0.49 & 0.70 & 0.54 & 0.46& 0.64 & 0.27 & 0.23 & 0.49 & 0.63 & 0.44 & 0.44  & 0.59 \\ 
AZ-NAS & \textbf{0.91} & \textbf{0.90} & \textbf{0.89} & 0.54 & 0.70 & 0.31 & 0.53 & 0.58 & \textbf{0.79}& 0.41& 0.60 & 0.72& 0.52& \textbf{0.65}& 0.90& 0.82& 0.77& \textbf{0.85}& 0.77\\ 
LIBRA (ours) & 0.89 & \textbf{0.90} & 0.87 & \textbf{0.77} & \textbf{0.74} & \textbf{0.45} & \textbf{0.57 }& \textbf{0.61} & \textbf{0.79} & \textbf{0.60} & \textbf{0.76} &\textbf{ 0.87} & \textbf{0.83} & 0.64 & \textbf{0.92} & \textbf{0.91} & \textbf{0.82} & \textbf{0.85} & \textbf{0.83} \\ 
\hline

\end{tabular}\vspace{-0.2cm}\caption{Spearman $\rho$ over different benchmarks on 1000 networks, obtained from multiple runs. All numbers were obtained in our experiments as in the original papers many experiments were run only for NB201, without specifing the \# test architectures, or directly to search the architecture on specific search-spaces reporting thus only the found test accuracy.}\vspace{-0.5cm}
\label{tab:libra_sota}
\end{table*}

\vspace{-0.1cm}
\subsection{Searching with LIBRA-NAS}\label{sec:exp_libra}
We now evaluate the performance of other ensembling methods and compare them with LIBRA.~As shown in~\cref{tab:libra_sota}, LIBRA outperforms other methods by a large margin in 13 out of 19 tasks.~In four tasks, it achieves comparable performance to the competitive AZ-NAS, while in the less informative NB201 search space, AZ-NAS slightly surpasses LIBRA on CIFAR-10 and ImageNet16-120.~We excluded the method introduced in~\cite{nasbenchsuitezero} from our comparison, as it requires training a predictor with 100 networks and therefore does not qualify as a pure ZC proxy method.~To search DNNs without training, we incorporate LIBRA into zero-shot search algorithms.~Specifically, we apply a pruning-based algorithm~\cite{chen2021nas} for the DARTS search space and an evolutionary algorithm for the Autoformer search space.~When deployed in the NAS search, LIBRA outperforms training-based methods while significantly reducing search time.~This is particularly evident on the more complex ImageNet task, where LIBRA identifies a network with 83\% test accuracy in just two hours, compared to CIFAR-10, where gains are smaller but still notable.
\begin{table}
\centering
\footnotesize
\renewcommand{\arraystretch}{1.2} 
\setlength{\tabcolsep}{3pt}
\begin{tabular}{>{\arraybackslash}m{1.0cm} >{\centering\arraybackslash}m{1.7cm} >{\centering\arraybackslash}m{1.1cm} >{\centering\arraybackslash}m{0.8cm} >{\centering\arraybackslash}m{1.5cm} >{\centering\arraybackslash}>{\columncolor{lightgray}}m{1.1cm}}
\hline
$\mathcal{B}_{ij}$ & NAS Method & Search approach & Params (M) & Search Time (GPU days) & Test Error (\%)\\
\hline\hline
\multirow{7}{*}{\parbox{1.3cm}{DARTS \\ Cifar-10}} & PC-DARTS & gradient & 3.6 & 0.1 & 2.57  \\
& AmoebaNet-A & evolution & 3.2& 3150 & 3.34 \\
& ENAS & RL & 4.6 & 0.5 & 2.89 \\
& SynFlow & TF & 5.08 & 0.11 & 7.85 \\
& AZ-NAS & TF & 4.1 & 0.4 & 2.55 \\
& SWAG & TF & 3.6 & 0.01 & 2.47 \\
& LIBRA & TF & 3.1 & 0.08 & 2.45 \\
\hline
\multirow{7}{*}{\parbox{1.3cm}{DARTS \\ IMNET1k}} & PC-DARTS & gradient &5.3 & 3.8 & 24.2 \\
& AmoebaNet-C & evolution & 6.4 & 3150 & 24.3\\
& NASNet-A & RL & 5.3 & 2000 & 26.0 \\
& SynFlow & TF & 6.3 & 0.5 & 30.1 \\
& AZ-NAS & TF & 6.2  & 0.7 & 23.6 \\
& SWAG & TF & 5.8 & 0.11 & 23.4 \\
& LIBRA & TF & 5.7 & 0.3 & 23.1 \\
\hline
\multirow{5}{*}{\parbox{1.3cm}{AutoFormer \\ Small \\ IMNET1k}} & Autoformer & evolution & 22.9 & 24 & 18.3 \\
& AZ-NAS & TF & 23.8 & 0.07 & 17.8 \\
& TF-TAS & TF & 23.9 & 0.5 & 18.1 \\
& SWAG & TF & 23.7 & 0.05 & 17.8 \\
& LIBRA & TF & 23.1 & 0.1 & 17.0 \\
\hline
\end{tabular}
\caption{Search results in DARTS and Autoformer searh space. TF = training free, RL = reinforcement learning, $\mathcal{B}_{ij}$ = benchmark for search-space $i$ in dataset $j$.}\vspace{-0.8cm}
\label{tab:libra_swag_nas}
\end{table}

\vspace{-0.1cm}
\subsection{Ablation}\label{sec:ablation}
\textbf{Influence of each L-SWAG component. }In~\cref{tab:ablation_swag} we ablate every component on a variety of search-spaces. We did not limited the ablation on NB201, as each component of L-SWAG has a different impact strength depending on the considered benchmark. 
For example, the first block, which analyzes each component independently, highlights that removing the mean has a stronger impact on TNB101’s Micro and Macro search spaces. Meanwhile, considering an interval of layers and including the expressivity term significantly affects TNB101 Macro, with a smaller impact on TNB101 Micro.~
Comparing the last rows of the 1$^{st}$ and 2$^{nd}$ blocks, we can observe how layer selection also improves consistently $\Psi$'s correlation across all search spaces.~Although NB201 is included for completeness, it provides limited insights aside from showing a steady gain when removing $\mu$ and selecting layers.~Across search spaces, a general trend emerges: choosing specific layers for gradient statistics has a strong positive effect on the Macro search space, while layer selection in the computation of $\Psi$ proves more beneficial for the Micro search space.
\begin{table}[ht]
    \centering
    \footnotesize
    \setlength{\tabcolsep}{3pt}
    \begin{tabular}{cccccccccccc}
        \toprule
        \multirow{2}{*}{no $\mu$} & \multirow{2}{*}{$\hat{L}$} & \multirow{2}{*}{$\Psi$} & \multicolumn{3}{c}{{\scriptsize{NB201}}}  & \multicolumn{3}{c}{{\scriptsize{Micro}}} & \multicolumn{3}{c}{{\scriptsize{Macro}}} \\
        
        \cmidrule(lr){4-6} \cmidrule(lr){7-9} \cmidrule(lr){10-12}
        & & & {\scriptsize{C10}} & \scriptsize{C100} & \scriptsize{In16-120} & \scriptsize{AE} & \scriptsize{Jig.} & \scriptsize{Norm.} & \scriptsize{AE} & \scriptsize{Jig.} & \scriptsize{Norm.}\\ 
        \midrule
        & & &  0.75 & 0.80 & 0.78 & 0.16 & 0.53 & 0.68 & 0.19 & 0.05 & 0.53 \\
        \cmark & & & \textbf{0.78} & \textbf{0.81} & \textbf{0.79} & \textbf{0.19 }& \textbf{0.54} & 0.68 & 0.20 & 0.40 & 0.64\\    
        & \cmark & & 0.77 & \textbf{0.81} & \textbf{0.79} & 0.18 & 0.53 & \textbf{0.69} & 0.24 & 0.32 & 0.61\\
        & & \cmark & 0.71 & 0.75 & 0.71 & 0.01 & 0.38 & 0.53 & \textbf{0.71} & \textbf{0.74} & \textbf{0.79}\\
       \midrule
        \cmark & \cmark & &  \textbf{0.79} &  \textbf{0.82} & \textbf{0.80} & \textbf{0.28} & \textbf{0.56} & \textbf{0.73 }& 0.37 & 0.56 & 0.80 \\  
       
         \cmark &  & \cmark &  \textbf{0.79} & \textbf{0.82} & \textbf{0.80} & {0.27} & 0.55 & 0.71  & 0.74 & 0.75 & 0.81\\
          & \cmark & \cmark &\textbf{0.79} & 0.77 & 0.75 & 0.11 & 0.45 & 0.55 & \textbf{0.76} & \textbf{0.76} & \textbf{0.82} \\
          \midrule
          \rowcolor{lightgray}
           \cmark & \cmark & \cmark  &  0.79 & 0.83 & 0.80 & 0.31 & 0.58 & 0.75 & 0.79 & 0.78 & 0.84\\  

        \bottomrule
    \end{tabular}\vspace{-0.1cm}
    \caption{Ablation study for each component of L-SWAG. The tick on ``no $\mu$" denotes not having the mean of gradients, which is the proof for the conclusion we drew with~\cref{theorem:3.1}, $\hat{L}$ ablates selecting percentiles, $\Psi$ ablates the expressivity term. The row with no ticks stands for $\log(\frac{\mu}{\sigma})$ for all layers up to depth L.}\label{tab:ablation_swag}\vspace{-0.3cm}
\end{table}
\cref{tab:ablation_swag} ablates the presence of the $\hat{L}$ found according to our method (\cref{sec:swag}), but we obviously ablated different values for $\hat{L}$. A visual summary is depicted in~\cref{fig:evo_rank}, which describes the evolution of the $\rho$ correlation depending on the selected percentile (\cf supp. material for full quantitative results). 

\textbf{LIBRA ablation study.} 
\cref{tab:ablation_LIBRA1} presents a comparison of methods for combining the first two metrics, while~\cref{tab:ablation_LIBRA2} evaluates the impact of adding a third metric, $z_3$, selected via bias matching.~Various approaches were tested for selecting $z_1$ and $z_2$, based on patterns observed in~\cref{fig:corr_mat1000}.~For instance, using gradient-free ZC proxies yields a clear advantage on TNB101-Macro, whereas gradient-based metrics perform slightly better on TNB101-Micro. We assessed whether categorizing ZC proxies by type produced larger gains compared to minimizing \textbf{IG}~\cref{eq:info_gain}.~Additionally, we compared these strategies with \textbf{IG} maximization and random selection.~Selecting $z_2$ according to the LIBRA strategy consistently outperformed other methods, with the performance margin varying by benchmark.~For NB301, where no specific metric type is favored, this margin is notably larger, while it narrows in search spaces that favor either gradient-free or gradient-based proxies.~Lastly, we tested methods for selecting $z_3$, finding bias matching to be the most effective, followed by bias minimization.
\vspace{-0.2cm}
\begin{table}[ht]
    \centering
    \scriptsize
    \renewcommand{\arraystretch}{1.2} 
    \setlength{\tabcolsep}{3pt}
    \begin{subtable}{0.48\textwidth} 
        \centering
        \begin{tabular}{>{\arraybackslash}m{1.35cm} >{\centering\arraybackslash}m{0.6cm} >{\centering\arraybackslash}m{0.7cm} >{\centering\arraybackslash}m{1.3cm} >{\centering\arraybackslash}m{0.8cm} >{\centering\arraybackslash}m{1.0cm} >{\centering\arraybackslash}m{1.0cm}}
            \hline
            $\mathcal{B}_{ij}$ & 2 $\nabla$ free & 2 $\nabla$ based & $\nabla$ free + $\nabla$ based & 2 random & best + $\max IG$ & best + $\min IG$ \\
            \hline\hline
            NB201$_{\text{In16-120}}$ & 0.77 & 0.80 & 0.86 & 0.62 & 0.64 & 0.86 \\
            NB301$_{\text{C10}}$ & 0.63 & 0.53 & 0.56 & 0.57 & 0.53 & 0.71 \\
            Micro$_{\text{scene}}$ & 0.73 & 0.73 & 0.74 & 0.41 & 0.62 & 0.77  \\
            Macro$_{\text{scene}}$ & 0.89 & 0.15 & 0.22 & 0.45 & 0.60 & 0.90 \\
            \bottomrule
        \end{tabular}
        \caption{1$^{st}$ column combines 2 best gradient free metrics, the 2$^{nd}$ two best gradient based, 3$^{rd}$ a gradient based and a gradient free with high $\rho$ on the $\mathcal{B}_{i,j}$. 4$^{th}$ random samples two metrics. Details on the selected proxies are provided in the supp. material.}
        \label{tab:ablation_LIBRA1}
    \end{subtable}
    \hfill
    \begin{subtable}{0.48\textwidth} 
        \centering
        \begin{tabular}{>{\arraybackslash}m{1.35cm} >{\centering\arraybackslash}m{1.0cm} >{\centering\arraybackslash}m{1.4cm} >{\centering\arraybackslash}m{1.9cm} >{\centering\arraybackslash}m{1.4cm} }
            \hline
            & w/o $b$ & w/ random $z_3$ & w/ $b$ minimization & w/ $b$ matching  \\
            \hline\hline
            NB201$_{\text{In16-120}}$ & 0.86 & 0.85 & 0.85 & 0.87\\
            NB301$_{\text{C10}}$ & 0.71 & 0.44 & 0.71 & 0.74\\
            Micro$_{\text{scene}}$ & 0.77 & 0.72 & 0.79 & 0.79\\
            Macro$_{\text{scene}}$ & 0.90 & 0.20 & 0.87 & 0.91\\
            \bottomrule
        \end{tabular}
        \caption{Ablations on the inclusion of the bias. The 2$^{nd}$ column chooses $z_3$ randomly, 3$^{rd}$ column chooses $z_3$ among well-performing ones, and minimizes its bias, 4$^{th}$ column, deployed in LIBRA, selects $z_3$ accoding to~\cref{algo:LIBRA}.}\vspace{-0.2cm}
        \label{tab:ablation_LIBRA2}
    \end{subtable}
    \caption{LIBRA component ablations.}
    \label{tab:ablation_LIBRA}\vspace{-0.5cm}
\end{table}

%% file: sec/5_conclusions.tex
\section{Conclusions}\label{sec:conclusions}
We proposed L-SWAG, a new ZC-proxy capturing expressivity and trainability of DNNs for ConvNets \emph{and} ViT, and LIBRA-NAS, a new ensemble algorithm to properly combine proxy metrics on a given benchmark. To this aim, we built a new benchmark composed of 2000 trained ViT models on six different tasks, and adapted previously introduced SOTA metrics to properly work on GeLU networks. To motivate the need of L-SWAG we evaluated all previously introduced ZC-proxies, under the same setup, on all benchmarks including our new Autoformer search space. We showed how L-SWAG achieves the best ranking consistency across several benchmarks. To motivate the need of LIBRA-NAS, we compared with other ML metric-aggregation methods and integrated LIBRA in the NAS search. In just 0.1 GPU days, LIBRA finds an architecture with a 17.0 \% test error on ImageNet1k, outperforming evolution and gradient-based NAS competitors.

\textbf{Limitations and Future work. }Our work makes progress towards expanding ZC-proxies to the ViT search space and toward providing a ML algorithm for combination of proxies. However, there are still some limitations.  First, our LIBRA evaluation is limited to an empirical analysis. Second, future work may extend L-SWAG to work on the video domain and for different input modalities. 
\section*{Acknowledgement}
This work has been partially supported by the project IN2814 of Free University of Bozen-Bolzano, by the Spanish project PID2022-136436NB-I00 and by ICREA under the ICREA Academia programme.

%% file: sec/X_suppl.tex
\clearpage
\appendix
\maketitlesupplementary

\section{Proof of Theorem~1}
\setcounter{theorem}{0}
\begin{theorem}
    Given a linear regressor $f(\va, \vx)$ with trainable parameters $\va = (a_j)_{j = 1}^M$, let $g(\vx_i) = (g_j(\vx_i))_{j = 1}^d$ be the gradient of $\va$ w.r.t. to $\vx_i$, and $\hat{\va} = \va -\eta\sum_i g_j(\vx_i)$ the updated parameters with learning rate $\eta$. Denote $\mu_j = \frac{1}{M}\sum_i g_j(\vx_i)$, $\sigma_j = \sqrt{\sum_i (g_j(\vx_i) - \mu_j)^2}$. Then, for any $\eta$, the total training loss $\mathcal{L}_f(\mX, \vy; \hat{\va}) = \frac{1}{2}\sum_i (\hat{\va}^\top \vx_i - y_i)^2$ of $f$ is bounded by:
    \begin{equation}
         \mathcal{L}_f(\mX, \vy; \hat{\va}) \le \frac{1}{2}\left(M\sum_{j = 1}^d \left[\sigma_j^2 + ((M\eta - 1)\mu_j)^2\right]\right).
    \end{equation}
\end{theorem}
\emph{Note. } There is an error in the proof of Theorem~3.1, in~\cite{Li2023Zico}. Going from the fourth to the fifth line of Eq.~23, the sum over $i$ on the third term is missing and it should, instead, be $\sum_{ij} \eta^2 M^2 \mu_j^2$. Additionally, the $1/2$ factor does not multiply all terms, when instead it should. We thus provide the correct proof for the theorem with a resulting corrected upper bound:
\begin{proof}
Given a training sample $(\vx_i, y_i)$, with $||\vx_i|| = 1$, the gradient of the MSE-based loss function $\mathcal{L}$ defined in~Eq. 3 w.r.t. $\va$ when taking $(\vx_i, y_i)$ as input is:
\begin{equation}
    g(\vx_i) = \frac{\partial \mathcal{L}(y_i, f(\vx_i; \va))}{\partial \va} = \vx_i \vx_i^\top \va - y_i \vx_i \tag{11}
\end{equation}

We note that:
\begin{align}
    (\va - g(\vx_i))^\top \vx_i - y_i &= \va^\top \vx_i - \va^\top \vx_i \vx_i^\top \vx_i + y_i \vx_i^\top \vx_i - y_i \nonumber \\
    &= \va^\top \vx_i - (\va^\top \vx_i)(\vx_i^\top \vx_i) \nonumber \\
    &= \va^\top \vx_i - \va^\top \vx_i \nonumber \\
    &= 0 \implies y_i = (\va - g(\vx_i))^\top \vx_i \tag{12}
\end{align}

Then the total training loss among all training samples is given by:
\begin{equation}
    \sum_{i=1}^M \frac{1}{2} \left( \hat{\va}^\top \vx_i - y_i \right)^2 \tag{13}
\end{equation}

By using Eq.~12, we can rewrite Eq.~13 as follows:
\begin{align}
    \sum_{i=1}^M \frac{1}{2} \left( \hat{\va}^\top \vx_i - y_i \right)^2 &= \sum_{i=1}^M \frac{1}{2} \left( \hat{\va}^\top \vx_i - (\va - g(\vx_i))^\top \vx_i \right)^2 \nonumber \\
    &= \sum_{i=1}^M \frac{1}{2} \left( (\hat{\va} - \va + g(\vx_i))^\top \vx_i \right)^2 \tag{14}
\end{align}

Recall the assumption that $\hat{\va} = \va - \eta \sum_i g(\vx_i)$; we rewrite Eq.~14 as follows:
{\small
\begin{equation}
    \sum_{i=1}^M \frac{1}{2} \left( \hat{\va}^\top \vx_i - y_i \right)^2 = \sum_{i=1}^M \frac{1}{2} \left(\left( g(\vx_i) - \eta \sum_i g(\vx_i) \right)^\top \vx_i \right)^2 \tag{15}
\end{equation}
}
According to the Cauchy–Schwarz inequality and $\|\mathbf{x}_i\| = 1$, the total training loss is bounded by:
{\small
\begin{align*}
    &\sum_{i=1}^M \frac{1}{2} \left( \hat{{\va}}^\top \vx_i - y_i \right)^2 \leq \\
    &\leq \frac{1}{2} \sum_{i=1}^M \| (g({\vx}_i) - \eta \sum_i g({\vx}_i)) \|^2 \cdot \|\vx_i\|^2 \\
    &= \frac{1}{2} \sum_{i=1}^M \| (g(\vx_i) - \eta \sum_i g(\vx_i)) \|^2 \\
    &= \frac{1}{2}\sum_{i=1}^M\sum_{j=1}^d (g_j(\vx_i) - \eta M \mu_j)^2 \\
    &= \frac{1}{2}\sum_{i=1}^M\sum_{j=1}^d \left( [g_j(\vx_i)]^2 - 2 \eta M \mu_j g_j(\vx_i) + \eta^2 M^2 \mu_j^2 \right) \tag{16} \\   
    &= \frac{1}{2} \left(\sum_{ij}[g_j(\vx_i)]^2 + \eta^2 M^3\sum_{j} \mu_j^2 - 2\eta\frac{1}{M} \sum_{ij} (M^2 \mu_j g_j(\vx_i)) \right) \\
    &= \frac{1}{2}\left(\sum_{ij}[g_j(\vx_i)]^2 - 2 \eta M^2 \sum_j\mu_j^2 + \eta^2 M^3 \sum_j\mu_j^2 \right) \\
    &= \frac{G}{2} - \frac{\eta}{2} M^2(2 - \eta M)\sum_j \mu_j^2.
\end{align*}
}

As $G = \sum_j\sum_i [g_j(x_i)]^2 = \sum_j (M\mu_j^2 + M\sigma_j^2)$. Then we can rewrite:
\begin{align}
  &\min_\va\sum_i\mathcal{L}(y_i, f(\vx_i; \va)) \le \notag \\
  &\frac{1}{2} M \sum_j \left( \sigma_j^2 + (M^2\eta^2 - 2M\eta + 1)\mu_j^2 \right).\tag{17}
\end{align}

This term is non-negative for all $\eta$, therefore it decreseases by decreasing $\mu_j$ and $\sigma_j$, for any $j$. Please note that for $\eta = \frac{1}{M}$ our bound reduces to Eq. 6 of ZiCO. 
\end{proof}
This result is supported by Fig.~\ref{fig:toy}. Following~\cite{Li2023Zico} we built the same experiment setup: we randomly sample 1000 training images from MNIST dataset and normalize them with their L2-norm to create the training set $\mathbb{S}$. With a batch of 1, we train the network for one epoch, compute the gradient w.r.t the network parameters for each individual sample, and update the weights with a learning rate $\eta=\{\frac{1}{10M}; 1; \frac{3}{M}\}$ for three different experiments to provide evidence that our result is valid for a range of $\eta$. We compute the square sum of mean gradients ($x$-axis in the plot) and the total loss $(y$-axis in the plot). We repeat the process 1000 times on the same $\mathbb{S}$, each time by randomly sampling a different initialization strategy among Kaiming Uniform, Kaiming Normal, Xavier Uniform, and Xavier Normal. The plots show a clear positive correlation for the linear network among the square sum of mean gradients and the loss, as supported by our bound. 

\setcounter{figure}{4}
\begin{figure}[t]
  \centering
  \begin{subfigure}{0.50\linewidth}
    \includegraphics[width=\textwidth]{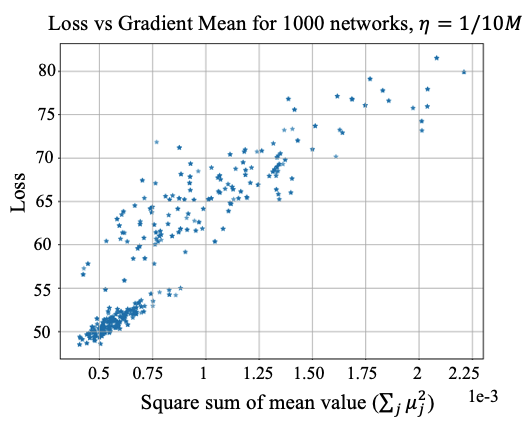}
    \caption{$\eta = \frac{1}{10M}$}
    \label{fig:toy1}
  \end{subfigure}
  \hfill
  \begin{subfigure}{0.48\linewidth}
    \includegraphics[width=\textwidth]{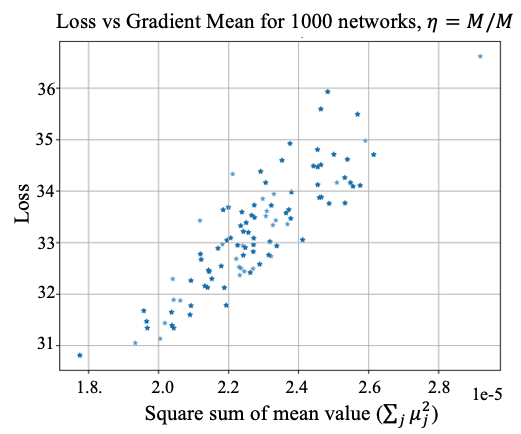}
    \caption{$\eta = 1$}
    \label{fig:toy2}
  \end{subfigure}
    \begin{subfigure}{0.49\linewidth}
    \includegraphics[width=\textwidth]{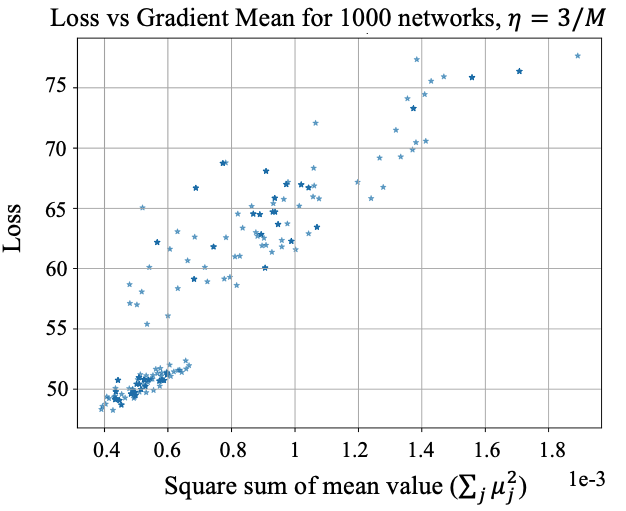}
    \caption{$\eta = \frac{3}{M}$}
    \label{fig:toy3}
  \end{subfigure}
  \caption{Toy example for the positive correlation of $\mu$ and the loss $\mathcal{L}$ for 1000 linear networks trained for one epoch on M = 1000 samples with different $\eta$. }\vspace{-0.5cm}
  \label{fig:toy}
\end{figure}

\section{Overview of the benchmarks}\label{sec:benchmark_overview}

\begin{figure*}[ht]
    \centering
    \begin{subfigure}[b]{\textwidth}
        \centering
        \includegraphics[width=0.8\textwidth]{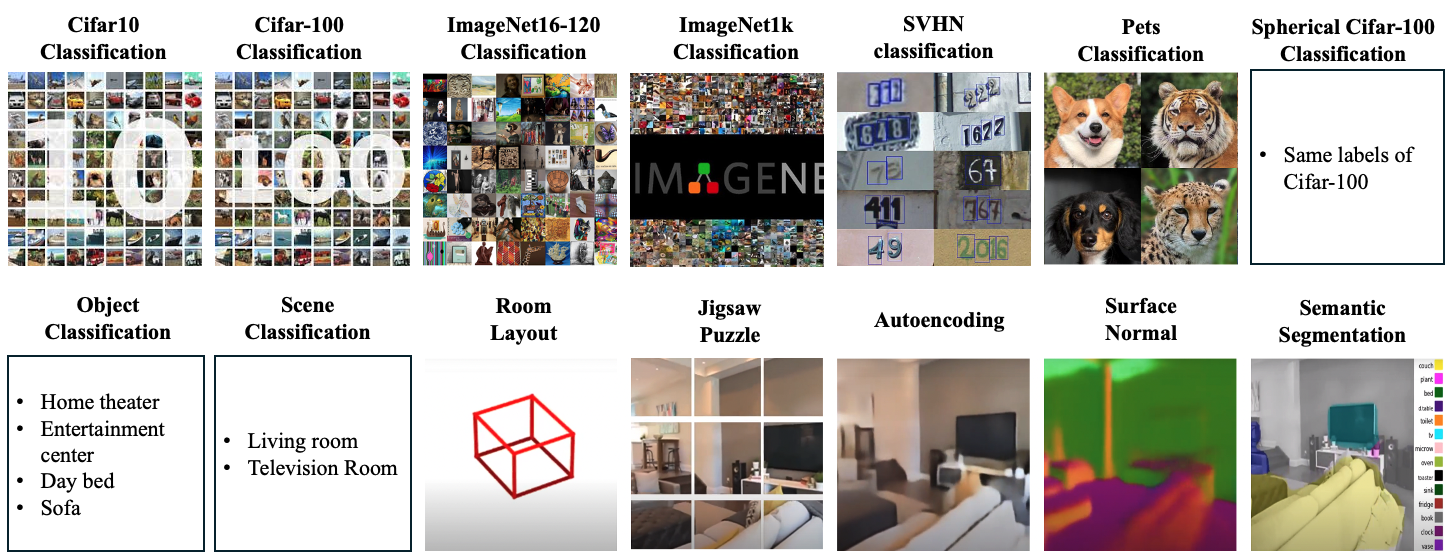}
        \caption{}
        \label{fig:details_benchmark_input}
    \end{subfigure}
    
    
    \begin{subfigure}[b]{\textwidth}
        \centering
        \includegraphics[width=0.81\textwidth]{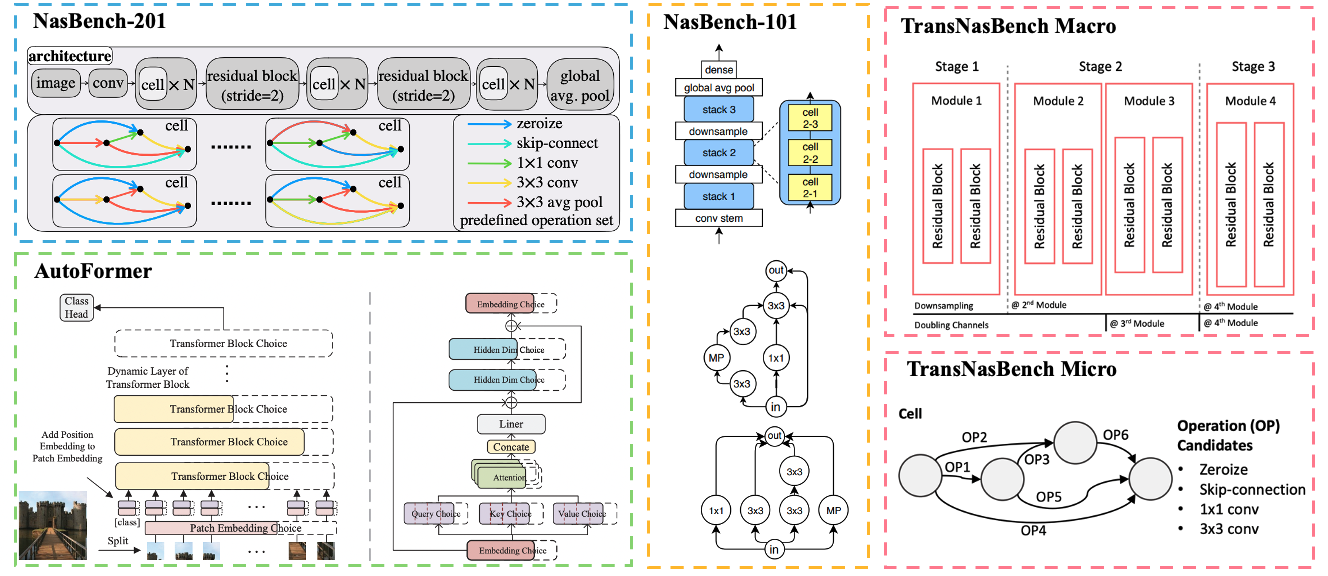}
        \caption{}
        \label{fig:details_benchmark_searchspace}
    \end{subfigure}\vspace{-0.2cm}
    \caption{Overview of the deployed datasets (\cref{fig:details_benchmark_input}) and search spaces (\cref{fig:details_benchmark_searchspace}) utilized in our work. We borrow the search-space images from the original NAS benchmark papers~\cite{nasbench101, dong2020nasbench201, autoformer, tnb101}.}\vspace{-0.4cm}
    \label{fig:details_benchmark}
\end{figure*}

In our experiments, we evaluate the proxies over 14 different tasks and across 6 different search spaces (see~\cref{fig:details_benchmark}). NasBench-101~\cite{nasbench101} is a cell-based search space consisting of 423 624 architectures. The design is thought to include ResNet-like and Inception-like DNNs trained multiple times on Cifar-10. In our full evaluation (see~\ref{sec:full} for details) we sampled and ranked 10 000 architectures from this search space. NasBench-201~\cite{dong2020nasbench201} is a cell-based search space composed of 15 625 architectures (6 466 of which are non-isomorphic) trained on 3 different tasks: Cifar-10, Cifar-100 and ImageNet-16-120. In our full evaluation, we ranked all 15 625 architectures. NasBench-301~\cite{nasbench301} (which is not depicted in~\cref{fig:details_benchmark}) is a cell-based search space created as a surrogate NAS benchmark for the DARTS search-space. DARTS is therefore composed of normal and reduction cells for a total of 10$^{18}$ different architectures trained on Cifar-10. In our full evaluation, we ranked 11 221 architectures. TransNAS-Bench-101~\cite{tnb101} is composed of a ``Macro" (with 3256 architectures) and a ``Micro" (cell-based) (with 4096 architectures) search space. Both Macro and Micro architectures are trained over 7 different tasks taken from the Taskonomy dataset. In our evaluation, we ranked all the 3256 + 4096 architectures. 
Finally, Autoformer~\cite{autoformer} is a one-shot architecture search space for Vision Transformers. We sampled 2000 architectures from the ``Small" search-space definition with Embedding dimension (320, 448, 64), $Q-K-V$ dimensions (320, 448, 64), MLP Ratio (3, 4, 0.5), Head Number (5, 7, 1), and Depth Number (12, 14, 1). The tuples of the three values in parentheses represent the lowest, highest, and steps values. We trained the 2 000 architecture on Cifar-10, Cifar-100, ImageNet-1k, SVHN, Pets and Spherical-Cifar100 datasets. 
\subsection{Autoformer Training}
We trained the Autoformer-Small search-space on two A100 Gpus with 80GB of memory each. We followed the standard training procedure introduced in~\cite{autoformer} and trained the One-Shot super network on ImageNet-1k splitting the images in $16\times16$ patches. The training was repeated three times with the weight-entanglement strategy introduced in~\cite{autoformer}, each time with 500 epochs (with 20 warmup epochs), an AdamW optimizer, 1024 batch size, lr=1e-3, cosine scheduler, weight-decay=5e-2, 0.1 label smoothing and dropout of 0.1. We used the average of the three runs as a test accuracy. The super network has been subsequently fine-tuned on Cifar-10, Cifar-100, Pets, SVHN, and Scifar-100 following the standard DeiT strategy~\cite{deit}. The 2000 architectures were sampled from the super network after training and directly evaluated with no further fine-tuning on the target dataset. 

\section{Full search-space}\label{sec:full}
\begin{figure*}[t]
  \centering
    \includegraphics[width=\textwidth]{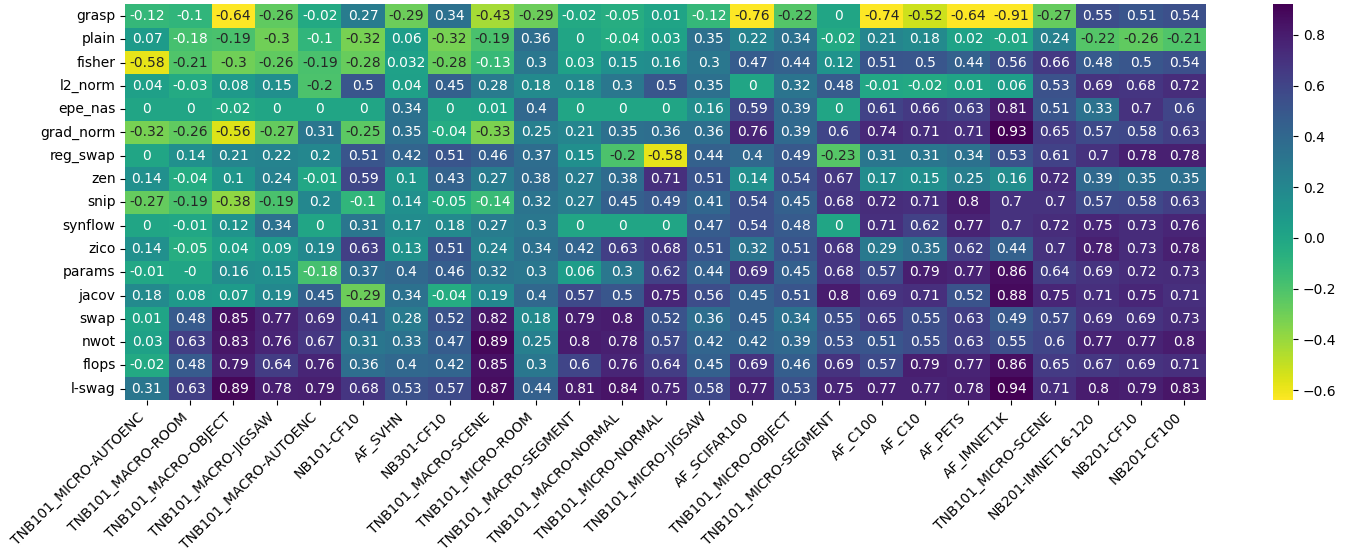}
    \caption{Spearman rank correlation coefficient between ZC proxy values and validation accuracies. Results were obtained from 5 multiple runs. Rows and columns are ordered based on the mean scores. This represents the results of Fig.~3 obtained for a larger number of architectures detailed in~\cref{sec:benchmark_overview}.}\vspace{-0.4cm}
    \label{fig:corr_matFULL}
\end{figure*}

\begin{figure*}[t]
    \centering
    \begin{subfigure}{0.24\textwidth}
        \centering
        \includegraphics[width=\linewidth]{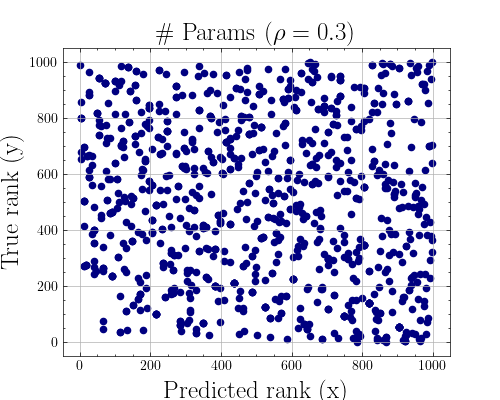}
        \caption{\# Params.}\label{fig:params_corr}
    \end{subfigure}
    \hfill
    \begin{subfigure}{0.24\textwidth}
        \centering
        \includegraphics[width=\linewidth]{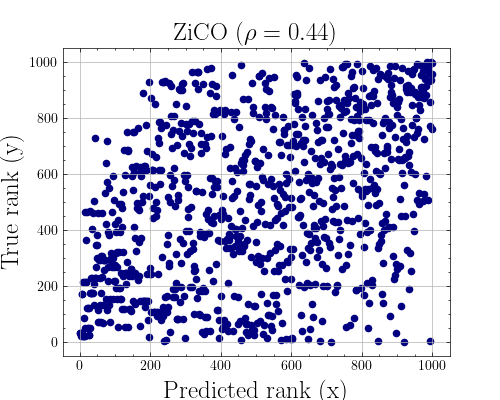}
        \caption{Zico.}\label{fig:zico_corr}
    \end{subfigure}
    \hfill
    \begin{subfigure}{0.24\textwidth}
        \centering
        \includegraphics[width=\linewidth]{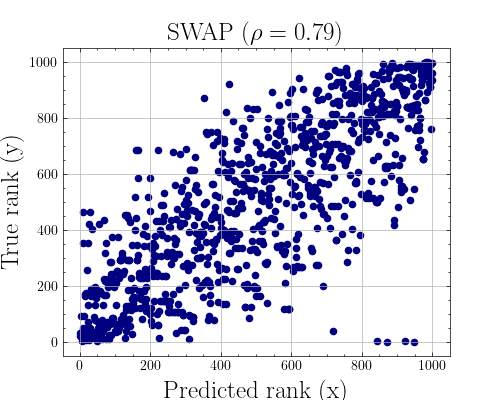}
        \caption{SWAP.}\label{fig:swap_corr}
    \end{subfigure}
    \hfill
    \begin{subfigure}{0.24\textwidth}
        \centering
        \includegraphics[width=\linewidth]{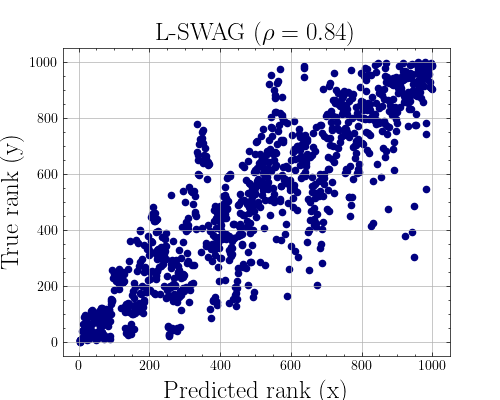}
        \caption{L-SWAG.}\label{fig:lswag_corr}
    \end{subfigure}
    \caption{Visual comparison of some ZC-proxy methods in terms of predicted ranking ($x-axis$) and validation accuracy ($y-axis$) on TransNasBench-101 Macro Normal search-space. Each figure reports the Spearman $\rho$ correlation coefficient.}\vspace{-0.4cm}
    \label{fig:correlation_example}
\end{figure*}

This section extends the experiments from Sec~4.1. For each benchmark and proxy we evaluated the Spearman $\rho$ correlation over a larger collection of architectures, \ie 10 000 fro Nasbench-101, 15 625 for NasBench-201, 11 221 for NasBench-301, 3256 for Tnb101-Macro, 4096 for Tnb101-Micro, and 2000 for Autoformer (see~\cref{fig:corr_matFULL}). Most metrics keep stable performance compared to Fig.~3 (that has the results for 1000 architectures), with slightly decreased values for SWAP and ZiCO and a large $\rho$ drop for reg-SWAP which now appears in the first half of the rows. We also present in~\cref{fig:correlation_example} a visual comparison between L-SWAG correlation, ZiCO~\cite{Li2023Zico}, SWAP~\cite{peng2024swapnas} and the simple metric \# parameters for TransNasBench-101 Macro Normal search-space. The plots display the predicted network rankings \emph{vs.} the ground-truth ranking for 1000 architectures. We compare L-SWAG against ZiCO and SWAP as they are the metrics most related to our contribution. We display the results for Macro Normal as it represents a challenging benchmark where the benefits of L-SWAG can be better appreciated.~\cref{fig:params_corr} and~\ref{fig:zico_corr} produce incorrect predictions frequently, leading to low-accuracy networks that are highly ranked and vice-versa. L-SWAG shows the strongest correlation with the ground truth visible through a reduced width across line $y=x$ compared to SWAP.  

\vspace{-0.2cm}
\section{Details from Section~3.1}
This section extends the layer-selection choice (\cref{sec:layer_choice}) with the complete set of plots for the gradient statistics behavior introduced in Sec.~3.1, quantitative results on the percentiles ablation depicted in Fig.~2a, and details of the gradient statistics across networks clustered by depth.~\cref{sec:moltiplication} details the choice of direct composing $\Lambda$ and $\Psi$ in Eq.~(1) through multiplication. 
\subsection{Layer-choice}\label{sec:layer_choice}
\begin{figure*}[ht]
    \centering
    \begin{subfigure}[b]{0.24\textwidth}
        \centering
        \includegraphics[width=\textwidth]{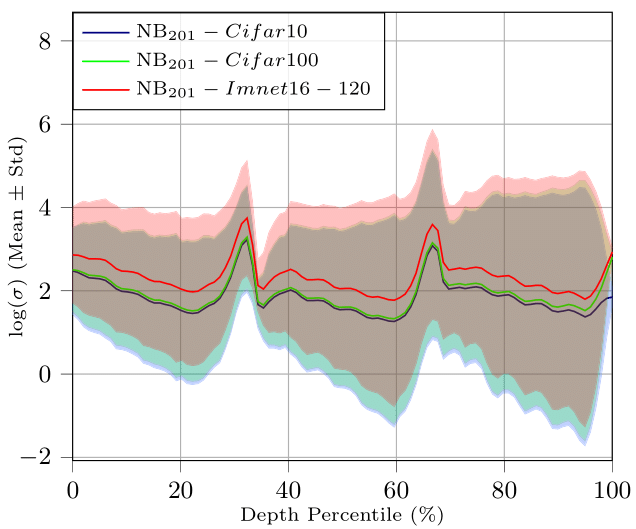}
        \caption{}
        \label{fig:nb201_layers}
    \end{subfigure}
    \hfill
    \begin{subfigure}[b]{0.24\textwidth}
        \centering
        \includegraphics[width=\textwidth]{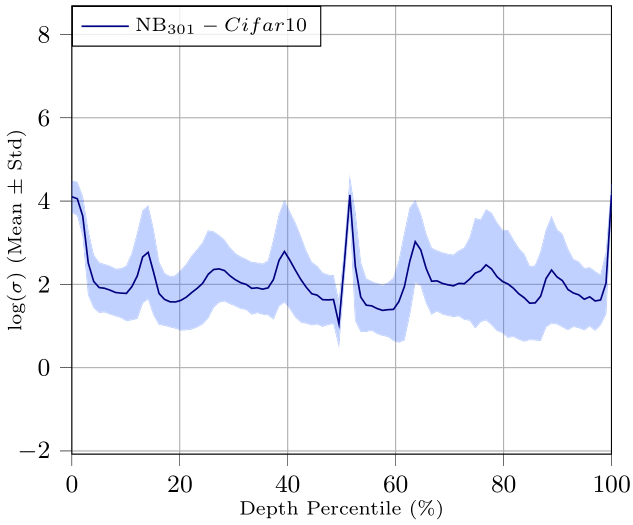}
        \caption{}
        \label{fig:nb301_layers}
    \end{subfigure}
    \hfill
    \begin{subfigure}[b]{0.24\textwidth}
        \centering
        \includegraphics[width=\textwidth]{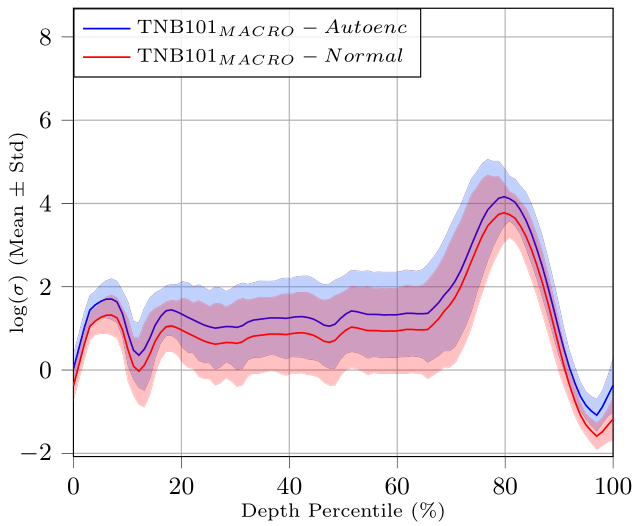}
        \caption{}
        \label{fig:tnb_macro_ae_norm}
    \end{subfigure}
    \hfill
        \begin{subfigure}[b]{0.24\textwidth}
        \centering
        \includegraphics[width=\textwidth]{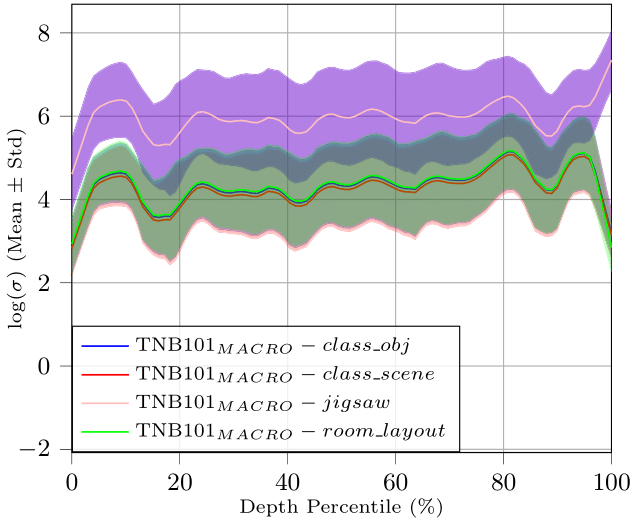}
        \caption{}
        \label{fig:tnb_macro_obj_scene_jig_room}
    \end{subfigure}
        \vspace{0.5cm}    
        \begin{subfigure}[b]{0.24\textwidth}
        \centering
        \includegraphics[width=\textwidth]{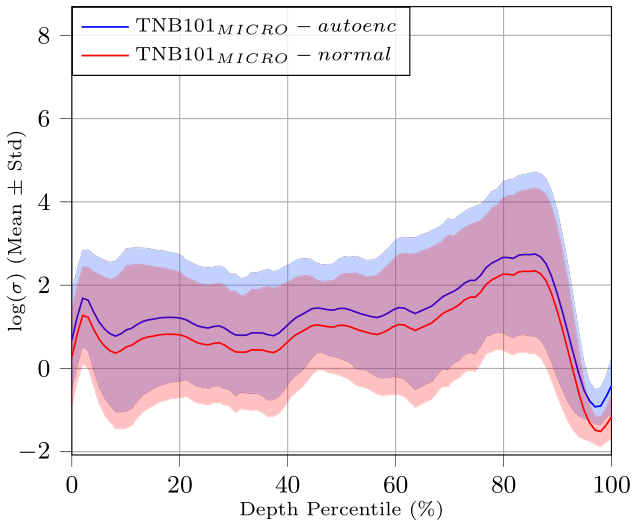}
        \caption{}
        \label{fig:tnb_micro_ae_norm}
    \end{subfigure}
    \hspace{0.1cm}
    \begin{subfigure}[b]{0.24\textwidth}
        \centering
        \includegraphics[width=\textwidth]{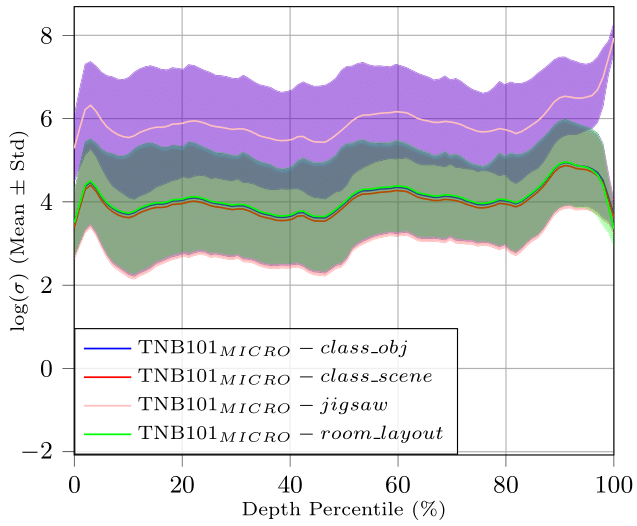}
        \caption{}
        \label{fig:tnb_micro_obj_scene_jig_room}
    \end{subfigure}
    \hspace{0.1cm}
    \begin{subfigure}[b]{0.24\textwidth}
        \centering
        \includegraphics[width=\textwidth]{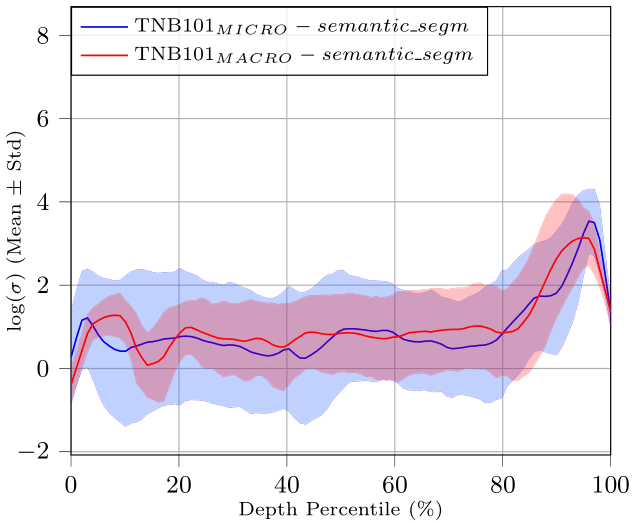}
        \caption{}
        \label{fig:tnb_sem_seg}
    \end{subfigure}\vspace{-0.8cm}
    
    \caption{Average gradient statistics across 1000 networks over different depth percentiles. This results completes Fig.~2 in the main paper.}
    \label{fig:SUP_full_layers_plot}
\end{figure*}

\begin{figure*}[ht]
    \centering
    \begin{subfigure}[b]{0.24\textwidth}
        \centering
        \includegraphics[width=\textwidth]{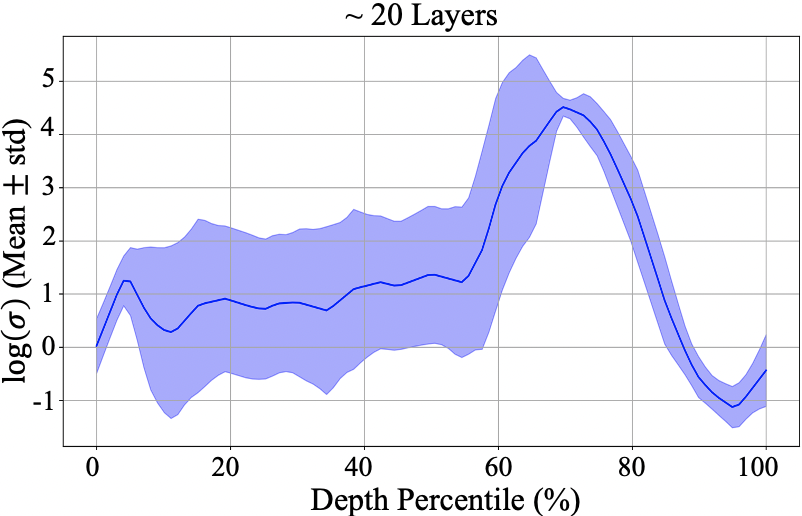}
        \caption{}
    \end{subfigure}
    \hfill
    \begin{subfigure}[b]{0.24\textwidth}
        \centering
        \includegraphics[width=\textwidth]{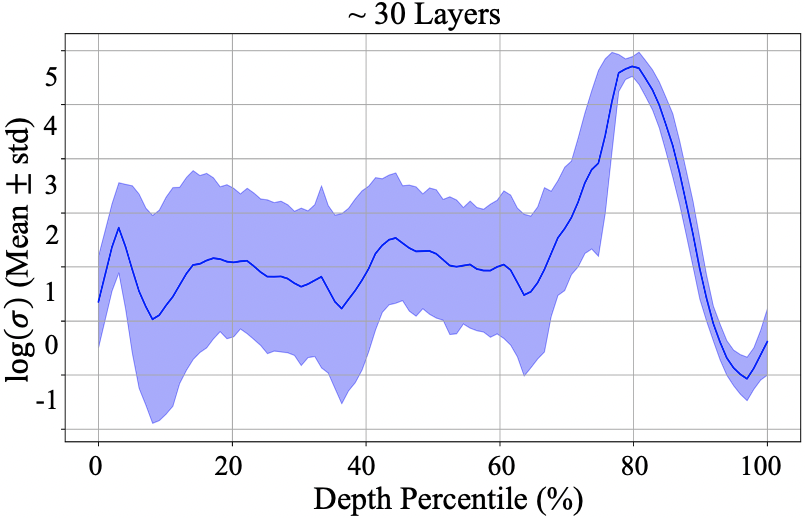}
        \caption{}
    \end{subfigure}
    \hfill
    \begin{subfigure}[b]{0.24\textwidth}
        \centering
        \includegraphics[width=\textwidth]{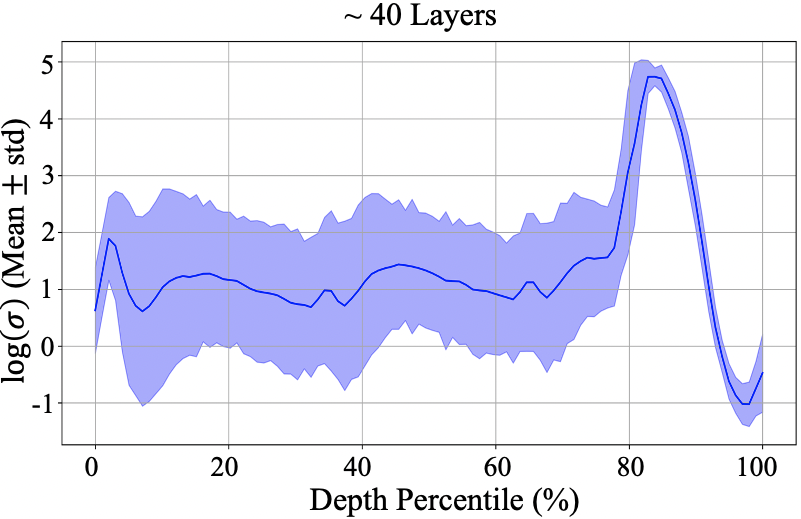}
        \caption{}
    \end{subfigure}
    \hfill
        \begin{subfigure}[b]{0.24\textwidth}
        \centering
        \includegraphics[width=\textwidth]{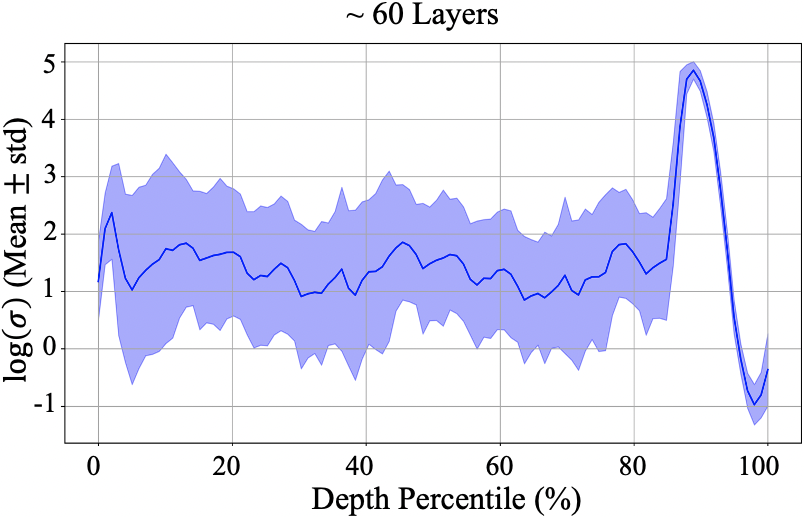}
        \caption{}
    \end{subfigure}
    
    \caption{Gradient statistics for different networks clustered by depth (20, 30, 40 and 60 layers) in TransBench101-Micro Autoencoder search space.}\vspace{-0.5cm}
    \label{fig:by_depth}
\end{figure*}

We organized the plots in~\cref{fig:SUP_full_layers_plot} by aggregating search-spaces with similar behavior. These graphs depict the mean and standard deviation of $\frac{1}{\Lambda}$ (introduced in Eq.~(1)) across 1000 randomly sampled networks. The goal is to highlight the intensity variation across different percentiles. The analysed search-spaces share different characteristics in the intensity trend, with~\cref{fig:nb301_layers} displaying NB301 periodic behavior,~\cref{fig:nb201_layers} highlighting three peaks (percentile 3, 7, and 10) in NB201,~\cref{fig:tnb_macro_ae_norm},~\cref{fig:tnb_micro_ae_norm} and~\cref{fig:tnb_sem_seg} presenting a unique peak shifted towards the last percentiles, and finally with~\cref{fig:tnb_micro_obj_scene_jig_room} and~\cref{fig:tnb_macro_obj_scene_jig_room} with an ascending intensity. If we couple these plots with the quantitative results in~\cref{tab:ablation_percentile} which ablates each percentile, and their visual representation in~Fig. {\color{cvprblue}2a} of the main paper, a clear match between the intensity of $\frac{1}{\Lambda}$ and the Spearman $\rho$ correlation emerges.~At this point, one may argue that the influence of the gradient statistics varies depending on the network depth, \ie we cannot average $\frac{1}{\Lambda}$ at the 8$^{th}$ percentile in a network with depth $L=100$ with $\frac{1}{\Lambda}$ at the 8$^{th}$ percentile in a network with depth $L=300$.~To clear any doubt, we show in~\cref{fig:by_depth} the same results of~\cref{fig:SUP_full_layers_plot} obtained by averaging only across networks with a comparable depth. We provide the example for Micro AutoEncoder search space as it represents the trend of all benchmarks.~Comparing~\cref{fig:tnb_micro_ae_norm} with~\cref{fig:by_depth} no substantial differences are observed. 


\setcounter{table}{4}
\begin{table*}[ht]
    \centering
    \scriptsize
    \setlength{\tabcolsep}{2pt}
   \begin{tabular}{> {\arraybackslash}m{0.9cm} >{\centering\arraybackslash}m{0.7cm} >{\centering\arraybackslash}m{0.7cm} >{\centering\arraybackslash}m{1.1cm}  >{\centering\arraybackslash}m{0.7cm} > {\centering\arraybackslash}m{0.7cm}  > {\centering\arraybackslash}m{0.7cm} >{\centering\arraybackslash}m{0.7cm} >{\centering\arraybackslash}m{0.7cm} >{\centering\arraybackslash}m{0.7cm} >{\centering\arraybackslash}m{0.7cm} >{\centering\arraybackslash}m{0.7cm} >{\centering\arraybackslash}m{0.7cm}  > {\centering\arraybackslash}m{0.7cm} >{\centering\arraybackslash}m{0.7cm} >{\centering\arraybackslash}m{0.7cm} >{\centering\arraybackslash}m{0.7cm} >{\centering\arraybackslash}m{0.7cm} >{\centering\arraybackslash}m{0.7cm} >{\centering\arraybackslash}m{0.7cm}}
    \hline

    \multicolumn{1}{c}{} & \multicolumn{3}{c}{{\scriptsize{NB201}}} & \multicolumn{1}{c}{\scriptsize{NB101}} &\multicolumn{1}{c}{\scriptsize{NB301}} & \multicolumn{7}{c}{\scriptsize{TNB101-Micro}} & \multicolumn{7}{c}{\scriptsize{TNB101-Macro}} \\
    \cmidrule(lr){2-4} \cmidrule(lr){5-5} \cmidrule(lr){6-6} \cmidrule(lr){7-13} \cmidrule(lr){14-20}
    \multirow{1}{*}{Percentile} & C10 & C100 & IN16-120& C10 &C10& AE & Room & Obj. & Scene & Jig. & Norm. & Segm. & AE & Room & Obj. & Scene & Jig. & Norm. & Segm.\\ 
        \midrule
       1  & 0.720 & 0.752 & 0.760 & 0.665 & \textbf{0.568} & \textbf{0.310} & 0.294 & 0.423 & 0.596 & 0.465 & 0.620 & 0.510 & \textbf{0.720}  & 0.030  & \textbf{0.890} & 0.745 & \textbf{0.780} & 0.700 & 0.650\\
       2  & 0.670 & 0.710 & 0.723 & 0.690 & 0.564 & 0.200 & \textbf{0.440} & \textbf{0.530} & \textbf{0.711} & \textbf{0.580} & \textbf{0.750} & 0.590 & 0.660  & \textbf{0.631}  & 0.700 & 0.830 & 0.720 & 0.800 & \textbf{0.810}    \\
       3  & 0.711 & 0.750 & 0.760 & \textbf{0.702} & 0.565 & 0.180 & 0.410 & 0.490 & 0.680 & 0.520 & 0.740 & 0.700 & 0.730  & 0.630  & 0.590 & 0.780 & 0.680 & \textbf{0.840} & 0.800   \\
       4  & 0.720 & 0.754 & 0.763 & 0.684 & 0.554 & 0.190 & 0.390 & 0.470 & 0.670 & 0.510 & 0.740 & 0.704 & 0.640  & 0.629  & 0.620 & 0.800 & 0.680 & 0.740 & \textbf{0.810}   \\
       5  & 0.690 & 0.730 & 0.743 & 0.687 & 0.549 & 0.240 & 0.420 & 0.468 & 0.690 & 0.540 & 0.730 & 0.690 & 0.580  & 0.628  & 0.670 & 0.810 & 0.690 & 0.610 & 0.740  \\
       6  & 0.720 & 0.751 & 0.759 & 0.680 & 0.554 & 0.170 & 0.390 & 0.480 & 0.680 & 0.520 & 0.720 & 0.695 & 0.620  & 0.628  & 0.690 & \textbf{0.870} & 0.710 & {0.670} & 0.740  \\
       7  & 0.720 & 0.751 & 0.760 & 0.690 & 0.550 & 0.110 & 0.390 & 0.470 & 0.670 & 0.510 & 0.630 & 0.690 & 0.530 & 0.625 & 0.540 & 0.750 & 0.730 & 0.630 & 0.710 \\
       8  & 0.655 & 0.690 & 0.710 & 0.685 & 0.540 & 0.060 & 0.420 & 0.490 & 0.690 & 0.530 & 0.530 & 0.690 & 0.550 & 0.625 & 0.590 & 0.760 & 0.630 & 0.630 & 0.660 \\
       9  & 0.717 & 0.749 & 0.757 & 0.682 & 0.540 & 0.080 & 0.380 & 0.460 & 0.660 & 0.500 & 0.510 & \textbf{0.750} & 0.520  & 0.627 & 0.600 & 0.760 & 0.700 & 0.550 & 0.710 \\
       10 & \textbf{0.724} &\textbf{ 0.760 }& \textbf{0.764} & 0.694 & 0.547 & 0.000 & 0.280 & 0.413 & 0.640 & 0.450 & 0.627 & 0.520 & 0.000  & 0.020  & 0.630 & 0.769 & 0.740 & 0.000 & 0.540\\
       
       \midrule

        ALL & 0.710 & 0.740 & 0.750 & 0.651 & 0.550 & 0.320 & 0.330 & 0.480 & 0.680 & 0.520 & {0.680} & 0.700 & 0.700 & 0.627 & 0.860 & 0.780 & 0.735 & 0.770 & 0.780\\
        \bottomrule
    \end{tabular}
    \caption{Collection of Spearman's $\rho$ correlation results obtained for the different percentiles. Each row represents an interval, \eg 1 refers to L-SWAG computed with $\hat{l}=0$ and $\hat{L}=1$, (meaning that for each row we calculated the metric with two percentiles). ``ALL" refers to the metric computed considering all the layers in a network. We highlight in bold the best results.  }\label{tab:ablation_percentile}\vspace{-0.5cm}
\end{table*}

\subsection{Multiplication}\label{sec:moltiplication}
In Eq.~(1) we directly combined $\Lambda$ and $\Psi$ through multiplication. As different metric combination strategies have been introduced in the literature, in this section we motivate such a choice. 
~\cite{chen2021nas} combined the ranks of architectures by averaging them across the constituent metrics, a strategy we refer to as ``RankAve". The advantage of RankAve lies in its equal weighting of contributions from each metric. However, this method also comes with several limitations. While rank aggregation is viable for certain search spaces and algorithms, it becomes impractical in many scenarios~\cite{lin2021zen}. Additionally, it is an indirect approach and arguably does not create a unified metric but instead offers a way to merge metrics. Similar to the method proposed in~\cite{tcet}, we consider addition and multiplication as alternative approaches. Consider two arbitrary metrics, $\tau_i$ and $\tau_j$, assumed to be independent random variables, where the samples represent evaluations of a network. For $k \in {i, j}$, we define $\mu_k = \mathbb{E}[\tau_k]$ and $\sigma_k^2 = \text{Var}(\tau_k)$. Starting with addition, we examine how to combine these metrics such that neither dominates the variance.
\[
\text{Var}(\tau_i + \tau_j) = \sigma_i^2 + \sigma_j^2.
\]

But what is the effect of the variance on the rankings? Suppose that $\sigma_i \gg \sigma_j$, then~\cite{ito1984probability}:
\[
P\big(|(\tau_i + \tau_j) - (\mu_i + \mu_j)| \geq k\big) \leq \frac{\sigma_i^2 + \sigma_j^2}{k^2} = \mathcal{O}(\sigma_i^2) 
\]
This suggests that the distributional characteristics of $\tau_i + \tau_j$ are primarily influenced by $\tau_i$, resulting in the overall ranking of architectures being controlled by $\tau_i$. Since it is improbable that the variances of the metrics are similar, the metric with the greater variance will dominate. Having excluded addition, we now proceed to evaluate multiplication:
\begin{align}
\text{Var}(\tau_i \cdot \tau_j) &= \sigma_i^2\sigma_j^2 + \mu_j^2\sigma_i^2 + \mu_i^2\sigma_j^2 + \mu_i^2\sigma_j^2 \notag \\
&+ \sigma_i^2\sigma_j^2 \bigg[1 + \bigg(\frac{\mu_j}{\sigma_j}\bigg)^2 + \bigg(\frac{\mu_i}{\sigma_i}\bigg)^2\bigg] \tag{19}
\end{align}

This highlights that the relationship between the metrics plays a more intricate role in determining the rankings. While not guaranteed, if the metrics' $\mu_k$ and $\sigma_k$ scale proportionally and exhibit similar distributional properties, this approach ensures that neither metric disproportionately dominates the variance. However, a legitimate concern arises: even when using metrics with minimal correlation, the assumption of independence may not always hold. Despite these limitations, we find evidence that, for $\Lambda$ and $\Psi$, the contributions of the individual components to the combined scores remain fairly balanced. Although more sophisticated operations than multiplication likely exist for direct composition, this analysis is intended solely as a proof of concept. An additional observation is that directly multiplying the final $\Lambda^{\hat{L}}$ and $\Psi$ results in the loss of much of the layer-wise information that has been gathered. This strengthens the case for our layer-wise multiplication via $\Psi^{\hat{L}}$, effectively performing a dot product of the layer-specific values. Such a layer-wise composition enables an assessment of individual layers based on their specific contribution to the network.

\section{Details from Sec.~3.2}
This section gives the details for the ZC-proxies $z_2, z_3$ that were chosen according to LIBRA algorithm and that provided the results of Tab~1. The first ZC-proxy $z_1$ can be simply derived from Fig.~3 by looking at each column (representing the benchmark) for the highest Spearman's $\rho$ correlation value. The metric that leads to the highest $\rho$ is selected as $z_1$. 
The second ZC-proxy $z_2$ is selected, according to Algorithm~1, by choosing among a filtered set of  ZC-proxies $z_h$. The $z_h$ with the lowest information gain $IG$ between $z_1$ and $z_h$ becomes $z_2$. The filtered set is obtained by discarding ZC-proxies with a Spearman's $\rho$ correlation below 0.1 points with respect to $z_1$ (for all cases otherwise specified). Following this rule, we selected $z_2 = jacov$ for NB201-C10, $z_2 = zico$ for NB201-C100, $z_2 = nwot$ for NB201-Imnet16-120, $z_2 = swap$ for NB301-C10, 
$z_2 = jacov$ for TNB101-micro-autoencoder, where the filtered set was obtaining with a tolerance of 0.2
$z_2 = epe-nas$ for TNB101-micro-room, 
$z_2 = l-swag$ for TNB101-micro-object, 
$z_2 = zen $ for TNB101-micro-scene, 
$z_2 = zico $ for TNB101-micro-jigsaw, 
$z_2 = jacov $ for TNB101-micro-normal, 
$z_2 = l-swag$ for TNB101-micro-segmentation, 
$z_2 = nwot $ for TNB101-macro-autoencoder,  
$z_2 = nwot$ for TNB101-macro-room,  where the filtered set was obtaining with a tolerance of 0.2
$z_2 = swap$ for TNB101-macro-object, 
$z_2 = swap $ for TNB101-macro-scene, 
$z_2 = swap $ for TNB101-macro-jigsaw, 
$z_2 = nwot $ for TNB101-macro-normal, 
$z_2 = nwot $ for TNB101-macro-segmentation. Although the choice in some cases (\eg Macro search-space) was restricted only to two/three ZC-proxies, as most of the $z_h$ had correlation below $\rho=0.4$, LIBRA could successfully identify the optimal choice. Let us take the example of TNB101-macro-jigsaw: the possible $z_h$ are {nwot with $\rho_{nwot}=0.76$, swap with $\rho_{swap}=0.74$, and flops with $\rho_{flops}=0.79$}. If we simply chose the metric with the highest $\rho$ (flops) we would obtain a $\rho_{z1,z2} = 0.79$, while LIBRA returns $\rho_{z1,z2} = 0.81$. Finally, in~\cref{tab:bias_values} we present the Pearson's correlation between all ZC-proxies and our chosen bias, \ie the number of parameters. We highlight in bold the ZC-proxy that was chosen according to LIBRA. 
\begin{table*}[ht]
    \centering
    \scriptsize
    \setlength{\tabcolsep}{2pt}
   \begin{tabular}{> {\arraybackslash}m{1.3cm} >{\centering\arraybackslash}m{0.6cm} >{\centering\arraybackslash}m{0.6cm} >{\centering\arraybackslash}m{1.1cm}  >{\centering\arraybackslash}m{0.6cm} > {\centering\arraybackslash}m{0.6cm}  > {\centering\arraybackslash}m{0.6cm} >{\centering\arraybackslash}m{0.6cm} >{\centering\arraybackslash}m{0.6cm} >{\centering\arraybackslash}m{0.6cm} >{\centering\arraybackslash}m{0.6cm} >{\centering\arraybackslash}m{0.6cm} >{\centering\arraybackslash}m{0.6cm}  > {\centering\arraybackslash}m{0.6cm} >{\centering\arraybackslash}m{0.6cm} >{\centering\arraybackslash}m{0.6cm} >{\centering\arraybackslash}m{0.6cm} >{\centering\arraybackslash}m{0.6cm} >{\centering\arraybackslash}m{0.6cm} >{\centering\arraybackslash}m{0.6cm}}
    \hline

    \multicolumn{1}{c}{} & \multicolumn{3}{c}{{\scriptsize{NB201}}} & \multicolumn{1}{c}{\scriptsize{NB101}} &\multicolumn{1}{c}{\scriptsize{NB301}} & \multicolumn{7}{c}{\scriptsize{TNB101-Micro}} & \multicolumn{7}{c}{\scriptsize{TNB101-Macro}} \\
    \cmidrule(lr){2-4} \cmidrule(lr){5-5} \cmidrule(lr){6-6} \cmidrule(lr){7-13} \cmidrule(lr){14-20}
    \multirow{1}{*}{Name} & C10 & C100 & IN16-120& C10 &C10& AE & Room & Obj. & Scene & Jig. & Norm. & Segm. & AE & Room & Obj. & Scene & Jig. & Norm. & Segm.\\ 
        \midrule
    \texttt{epe-nas} & 0.09 & 0.06 & 0.09 & -0.02  & 0.07& 0.43 & 0.25 & 0.22 & 0.30 & 0.17 & 0.40 & 0.32 & 0.13 & 0.12 & 0.10 & 0.11 & 0.02 & 0.12 & 0.26\\
    \texttt{fisher} & 0.16 & 0.15 & 0.07 &0.11 & 0.12 & 0.16 & 0.10 & 0.08 & 0.18 & 0.02  & 0.12 & 0.10 & 0.03 &  0.04 & 0.02 & 0.09 & 0.10 & 0.16 & 0.20 \\
    \texttt{flops} & 0.99 & 0.99 & 0.99 & 1.00 & 0.98 & 0.96 & 0.95 & 0.99 & 0.99 & 1.00 & 0.98 & 0.99 & 0.34 & 0.49 & 0.54 & \textbf{0.53} & 0.51 & 0.39 & 0.45\\
    \texttt{grad-norm} & 0.33 & 0.40 & 0.37 & 0.30 & 0.55 & 0.51 & 0.66 & 0.70 & 0.68 & 0.56 & 0.47 & 0.65 & 0.49 & 0.34 & 0.31 & 0.30 & 0.20 & 0.32 & 0.01 \\
    \texttt{grasp} & 0.05 & 0.03 & 0.13 & -0.03 & 0.16 & 0.18 & 0.12 & -0.20 & -0.23 & -0.35 & 0.20 & 0.15 &  0.08 & 0.16 & 0.11 & 0.06 & 0.08 & -0.21 & -0.04\\
    \texttt{l2-norm} & 0.69 & 0.69 & 0.69 & 0.62 & 0.99 & 0.64 & 0.17& 0.79 & 0.70 & 0.01& 0.64 & \textbf{0.51} & 0.49 & 0.24 & 0.76 & 0.45 &0.22 & 0.85 & 0.47\\
    \texttt{jacov} & 0.06 & 0.06 & 0.06 & -0.18 & 0.11 &\textbf{ 0.17} & 0.00& -0.03 & -0.00 &  \textbf{0.41} & 0.15 & 0.18 & 0.09 & 0.09 & 0.07  & 0.23 & 0.14 & 0.32  & \textbf{0.11}\\
    \texttt{nwot} & 0.51 & \textbf{0.51} & 0.50  & 0.74 & 0.95 & 0.42 & 0.35 & 0.46 & 0.40 & 0.35 & 0.07 & 0.35 & 0.19 & \textbf{0.24} & \textbf{0.31} & 0.30 & 0.21 & 0.34 & 0.00\\
    \texttt{params} & 1.00 & 1.00& 1.00 & 1.00 & 1.00 & 1.00& 1.00& 1.00 & 1.00& 1.00& 1.00& 1.00& 1.00& 1.00 &1.00& 1.00& 1.00 & 1.00 & 1.00\\
    \texttt{plain} & 0.32 & 0.10 & 0.23 & 0.03 & 0.39 & 0.12 & 0.15 & 0.05  & 0.08 & 0.50 & 0.19 & 0.10 & 0.09 & 0.08 & 0.17 & 0.11 & 0.34  & 0.06 & 0.49\\
    \texttt{snip} & 0.46 & 0.45 & 0.42 & 0.44 & 0.55 & 0.49 & 0.33 & 0.22 & 0.19 & 0.29& 0.55 & 0.21 & 0.39 & 0.28 & 0.55 & 0.45 & 0.49 & 0.68 & 0.53\\
    \texttt{synflow} & 0.24 & 0.24 & 0.24 & 0.57 & 0.07 & 0.41 & 0.05 & 0.45 & \textbf{0.40} & 0.44 & 0.47 & 0.11 & 0.27 & 0.12 & 0.23 & 0.21 & 0.35 & 0.41 & 0.23 \\
    \texttt{reg-swap} & 0.29 & 0.30 & 0.21 & 0.30 & \textbf{0.44} & -0.06 & 0.11 & -0.09 & 0.23 & -0.78 & 0.09 & -0.15 & 0.03 & -0.09 & 0.11 & -0.02 & 0.10 & 0.05 & 0.03 \\
    \texttt{zico} & 0.60 & 0.60 & \textbf{0.60} & \textbf{0.54} & 0.97 & 0.55& \textbf{0.48 }& \textbf{0.54}& 0.80  & 0.44 &\textbf{ 0.47} & 0.48 & 0.72 & 0.59 & 0.46 & 0.15 & 0.41 & 0.45 & 0.30 \\
    \texttt{swap} &  \textbf{0.50} & 0.51 & 0.47 & 0.44 & 0.50 & 0.01 & 0.35 & 0.30 & 0.35 & 0.21 & 0.35 & 0.29 & \textbf{0.32} & 0.41 & 0.54 & 0.12 & \textbf{0.11} & \textbf{0.39} & 0.36 \\
    \texttt{l-swag} & 0.23 & 0.24 & 0.24 & 0.19 & 0.32 & 0.00 & 0.08 & 0.15 & 0.19 & 0.17 & 0.15 & 0.21 & 0.02 & 0.16 & 0.18 & 0.22 & 0.10 & 0.21 & 0.11 \\
    \texttt{val-acc} & 0.41 & 0.55 & 0.57 & 0.47 & 0.52 & 0.18 & 0.40 & 0.53 & 0.54 & 0.43 & 0.44 & 0.59 & 0.05 & 0.07 & 0.24 & 0.41 & 0.16 & 0.40 & 0.08\\

        \bottomrule
    \end{tabular}
  \caption{Pearson correlation coefficients between predictors and our bias metric (\# of Parameters) on different benchmarks. We highlight in bold the value corresponding to the $z_3$ we chose for LIBRA. }
  \label{tab:bias_values}\vspace{-0.5cm}
\end{table*}

\section{Influence of the mini-batch size and of random initialization. }
\begin{figure*}[ht]
    \begin{subfigure}[b]{0.4\textwidth}
        \centering
        \includegraphics[width=\textwidth]{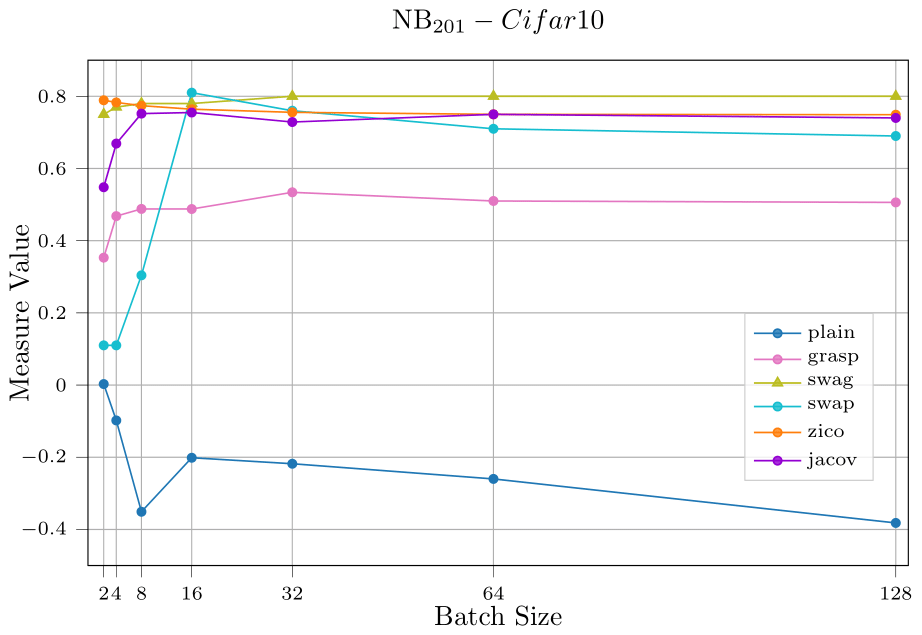}
        \caption{}
        \label{fig:ablation_batch_nb201}
    \end{subfigure}
    \hspace{2cm}
        \begin{subfigure}[b]{0.4\textwidth}
        \centering
        \includegraphics[width=\textwidth]{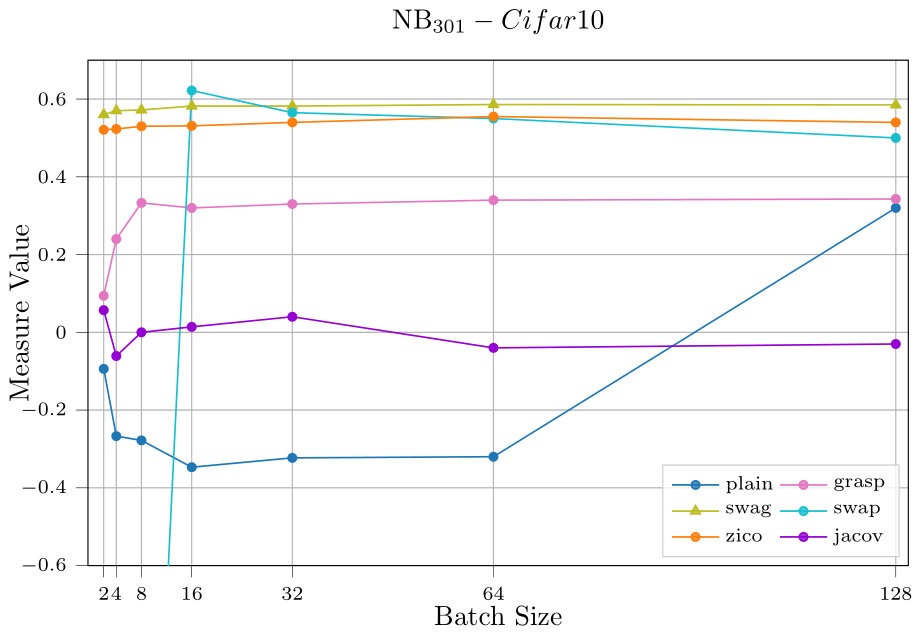}
        \caption{}
        \label{fig:ablation_batch_nb301}
    \end{subfigure}

    \begin{subfigure}[b]{0.4\textwidth}
        \centering
        \includegraphics[width=\textwidth]{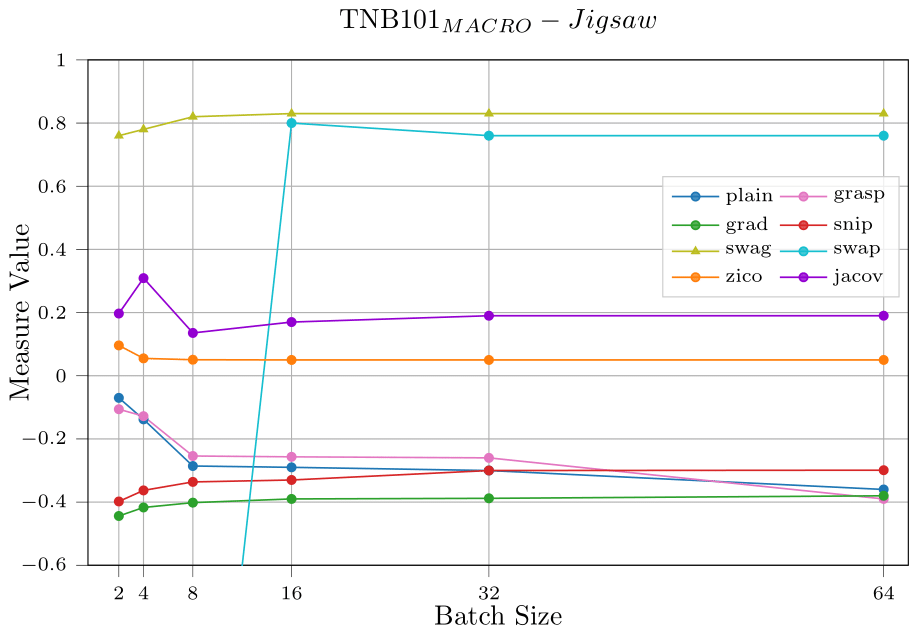}
        \caption{}
        \label{fig:ablation_batch_macrojigsaw}
    \end{subfigure}
    \hspace{2cm}
    \begin{subfigure}[b]{0.4\textwidth}
        \centering
        \includegraphics[width=\textwidth]{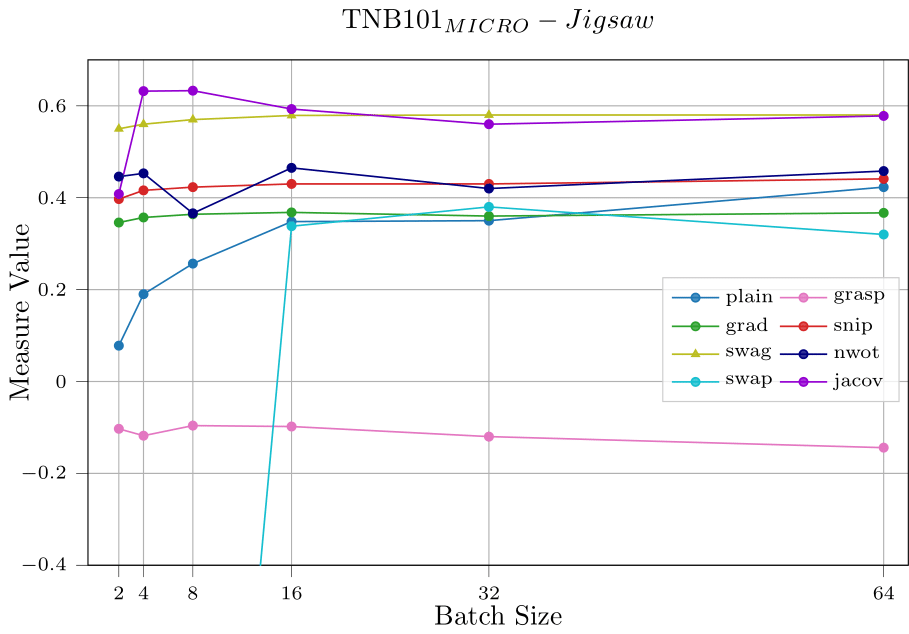}
        \caption{}
        \label{fig:ablation_batch_microjigsaw}
    \end{subfigure}
        \caption{Spearman $\rho$ coefficient consistency of ZC-proxies across different batch sizes.}
        \label{fig:ablation_batch}
\end{figure*}

We run ablation on the batch size for all measures, including our L-SWAG. We report a representative result for each search-space in~\cref{fig:ablation_batch}. Compared to the other measures in the plots, L-SWAG stabilizes after batch 32 saturating (differently from ZiCo which slightly deteriorates, or to SWAP which in fig.~\ref{fig:ablation_batch_nb201},~\ref{fig:ablation_batch_nb301} and~\ref{fig:ablation_batch_macrojigsaw} has its peak at B=16). We also noticed plain being highly unstable depending on the batch-size. Other metrics (\eg Fisher, \# flops \etc) with constant values across batches were simply not plotted. We also tested the measure with 3 different random initializations (Xavier, Kaiming and Gaussian) and found the metric to be robust with a std $\sigma=0.02$.\\
\vspace{-0.4cm}

\section{Information theory}
For the sake of clarity, we provide full details from Sec.~4.2 and provide the definition of Entropy borrowed from~\cite{nasbenchsuitezero}. 
Given two variables $y$ and $z_i$, the conditional entropy of $y$ given $z_i$ is defined as:
\begin{align}
    H(y | z_i) &= \mathds{E}[-\log(p(y|z_i))] \notag \\
               &= - \sum_{z \in \mathcal{Z}, y \in \mathcal{Y}} p(z, y) \log \frac{p(z, y)}{p(z)} 
     \label{eq:entropy}\tag{20}
\end{align}
for two support sets $\mathcal{Y, Z}$. If we consider entropy as a measure of information—or equivalently, the uncertainty associated with a random variable—conditional entropy reflects the remaining uncertainty after conditioning on another variable. Specifically, \( H(y \mid z_i) \) possesses several desirable properties: (1) \( H(y \mid z_i) = 0 \) if and only if \( z_i \) completely determines \( y \); (2) \( H(y \mid z_i) = H(y) \) if and only if \( y \) and \( z_i \) are entirely independent; and (3) \( H(y \mid z_{i1}, z_{i2}) = H(y, z_{i1}, z_{i2}) - H(z_{i1}, z_{i2}) \). This allows for straightforward computation of conditional entropy when conditioning on multiple random variables. Thus, it serves as an effective metric for quantifying remaining uncertainty or incomplete information. Following the above definition, would require all random variables to be discrete to compute the conditional entropy, which is not our case. Similarly to~\cite{nasbenchsuitezero}, to properly implement conditional entropy we use Sturge's rule~\cite{scott2009sturges} to discretize the float values describing $z_i$s. The heuristic to choose the number of bins is:\vspace{-0.2cm}
\begin{align}
    n_{bins} = \texttt{round(1+3.322*log(N))}, \notag \\ \text{with N = sample size.} \notag
\end{align}
Information about $y$ does not reveal the exact validation accuracy but rather the interval in which the value falls.

\begin{table*}[ht]
\vspace{0.2cm}
    \centering
    \renewcommand{\arraystretch}{1.0} 
    \setlength{\tabcolsep}{3pt}
    \begin{tabular}{>{\arraybackslash}m{1.4cm} >{\centering\arraybackslash}m{1.4cm} >{\centering\arraybackslash}m{1.5cm} >{\centering\arraybackslash}m{1.5cm} >{\centering\arraybackslash}m{1.5cm} >{\centering\arraybackslash}m{1.3cm} >{\centering\arraybackslash}m{1.7cm}> {\centering\arraybackslash}m{1.5cm} >{\centering\arraybackslash}m{1.5cm} >{\centering\arraybackslash}m{1.5cm}}
        \hline
        & Backbone & $AP^b$ & $AP^b_{50}$ & $AP^b_{75}$ & &$AP^M$ & $AP^M_{50}$ & $AP^M_{75}$  \\
        \hline
          
        & ResNet-50 & 38.0 & 58.6 & 41.4 & & 34.4 & 55.1 & 36.7 \\
        & PoolFormer & 40.1 & 62.2 & 43.4 & &37.0 & 59.1 & 36.9 \\
        \multirow{1}{*}{Object}& Swin-T & 43.7 & 66.6 & 47.7 &\multirow{1}{*}{Instance}  &39.8 & 63.3 & 42.7 \\
        \multirow{1}{*}{Detection} & $\epsilon$-GSNR & 45.0 & 67.1 & 49.1 & Segmentation &40.7 & 68.8 & 43.7 \\
        & L-SWAG & \textbf{47.5} & \textbf{71.4} & \textbf{50.3} & &\textbf{41.4} & \textbf{69.7} & \textbf{44.2} \\
        \bottomrule
    \end{tabular}\vspace{-0.2cm}
    \caption{{Comparison with models on COCO dataset.}}\vspace{0cm}
    \label{tab:obj_det_ist_seg}
\end{table*}

\section{LIBRA-NAS and L-SWAG-NAS: more results}
We extended the experiments presented in Sec.~4.2 for the Autoformer search space on ImageNet-1k. Rather than comparing with other training-free guided search methods, the focus of this set of experiments is to assess the benefit of ZC-NAS compared to other search methods deployed for the Autoformer search space, including simple random search. Although in~\cref{tab:libra_swag_MORE_AFnas} random search still represents a strong alternative, with the best-found architecture after three runs having a test error of 19 \%, both L-SWAG NAS and LIBRA-NAS largely improves performance of the found architecture with a negligible search-time. Given the large save in computation time, we hope this set of experiments will further convince the exploration of ZC-proxy design for the ViT search space, to expand research in the video domain. 
\begin{table}[h]
\centering
\footnotesize
\renewcommand{\arraystretch}{1.2} 
\setlength{\tabcolsep}{2pt}
\begin{tabular}{>{\arraybackslash}m{1.3cm} >{\centering\arraybackslash}m{2.5cm} >{\centering\arraybackslash}m{1.2cm} >{\centering\arraybackslash}m{1.5cm} >{\centering\arraybackslash}>{\columncolor{lightgray}}m{1.1cm}}
\hline
$\mathcal{B}_{ij}$ & Search approach & Params (M) & Search Time (GPU days) & Test Error (\%)\\
\hline\hline
\multirow{9}{*}{\parbox{1.3cm}{AutoFormer \\ Small \\ IMNET1k}}  & Weight entanglement + evolution & 22.9 & 24 & 18.3 \\
 & Random search & 23.0 & 0 & 19.0\\
& Classical weight sharing + random$\dagger$& 22.9 & - & 30.3 \\
 & Weight entanglement + random$\dagger$ &  22.8 & - & 18.7 \\
 & Classical weight sharing + evolution$\dagger$ & 22.9 & - & 28.5 \\
 & ViTAS~\cite{vitas}$\dagger$ & 30.5 & - & 18.0 \\
 & NASViT-A0\cite{nasvit}$\dagger$ & [200-300] & - & 21.8 \\
& L-SWAG-NAS  & 23.7 & 0.05 & 17.8 \\
& LIBRA-NAS & 23.1 & 0.1 & 17.0 \\
\hline
\end{tabular}
\caption{Further comparisons of networks from the Autoformer search space optimized by different NAS methods. While in Tab.~2 we mainly compared the search results obtained running the search algorithm guided by different ZC proxies evaluation, this set of experiments aims instead at showing the benefits of our contributions with respect to other NAS search methods. Random search is performed three times and the best performance is reported. $\dagger$Results were borrowed directly from~\cite{autoformer} and for such a reason no search time is reported, as not available in the original paper.}
\label{tab:libra_swag_MORE_AFnas}
\end{table}
We run additional experiments on BurgerFormer~\cite{burgerformer} for object detection and instance segmentation on COCO dataset and will add the following results in Sec.~H SM. We chose~\cite{burgerformer} to be comparable with $\epsilon$-GSNR which also validates the metric on these tasks. As $\epsilon$-GSNR, we deployed the found network from BurgerFormer-S space (pre-trained on ImageNet 83.5 \% acc.) as the backbone for the Mask R-CNN detector. We used an evolutionary algorithm to search networks within 30M Params.

\section{Theoretical intuition behind L-SWAG for ViT}
This brief section aims at delivering the intuition behind the design of L-SWAG and the motivation of why it works on ViTs.
ViTs use MSA to capture long-range dependencies, but a common issue is rank collapse, where MSA outputs converge to rank-1 matrices, reducing representational diversity. Activation patterns in MSA reflect self-attention’s ability to distinguish input tokens. Greater diversity in these patterns at initialization indicates higher expressivity, avoiding rank collapse~\cite{pmlr-v119-bhojanapalli20a}. While GELU is nonlinear, its smooth transitions still separate input space into ``soft regions", which can be counted like in ReLU.
Gradient variance ensures trainability, as GELU's smoothness can lead to gradient issues. 
Together, they provide a holistic measure of both expressivity and trainability.